\documentclass{article}

\usepackage{microtype}
\usepackage{graphicx}
\usepackage{subfigure}
\usepackage{booktabs} 

\usepackage{hyperref}

\usepackage[accepted]{icml2025}

\usepackage{amsmath}
\usepackage{amssymb}
\usepackage{mathtools}
\usepackage{amsthm}

\usepackage[capitalize,noabbrev]{cleveref}

\theoremstyle{plain}

\theoremstyle{definition}

\theoremstyle{remark}

\usepackage[textsize=tiny]{todonotes}

\usepackage{tcolorbox}
\usepackage{fancyvrb,fvextra}
\usepackage{enumitem}
\usepackage{booktabs}
\usepackage{multirow}
\usepackage{subfigure}
\usepackage[fixed]{fontawesome5}

\usepackage{makecell}

\newcommand{\ApproachName}{\textsc{IterIT}}

\icmltitlerunning{Boosting LLM via Learning from Data Iteratively and Selectively}

\begin{document}

\twocolumn[
\icmltitle{Boosting LLM via Learning from Data Iteratively and Selectively}



\icmlsetsymbol{equal}{*}

\begin{icmlauthorlist}

\icmlauthor{Qi Jia}{comp}
\icmlauthor{Siyu Ren}{yyy}
\icmlauthor{Ziheng Qin}{comp}
\icmlauthor{Fuzhao Xue}{dm}
\icmlauthor{Jinjie Ni}{comp}
\icmlauthor{Yang You}{comp}
\end{icmlauthorlist}

\icmlaffiliation{comp}{National University of Singapore}
\icmlaffiliation{yyy}{Meituan}
\icmlaffiliation{dm}{Work done in National University of Singapore, now in Google DeepMind}

\icmlcorrespondingauthor{Yang You}{youy@comp.nus.edu.sg}

\icmlkeywords{Machine Learning, ICML}

\vskip 0.3in
]



\printAffiliationsAndNotice{}  

\begin{abstract}

Datasets nowadays are generally constructed from multiple sources and using different synthetic techniques, making data de-noising and de-duplication crucial before being used for post-training. In this work, we propose to perform instruction tuning by iterative data selection (\ApproachName{}). We measure the quality of a sample from complexity and diversity simultaneously. 
Instead of calculating the complexity score once for all before fine-tuning, we highlight the importance of updating this model-specific score during fine-tuning to accurately accommodate the dynamic changes of the model. On the other hand, the diversity score is defined on top of the samples' responses under the consideration of their informativeness.
\ApproachName{} integrates the strengths of both worlds by iteratively updating the complexity score for the top-ranked samples and greedily selecting the ones with the highest complexity-diversity score. 
Experiments on multiple instruction-tuning data demonstrate consistent improvements of \ApproachName{} over strong baselines. 
Moreover, our approach also generalizes well to domain-specific scenarios and different backbone models. All resources will be available at \url{https://github.com/JiaQiSJTU/IterIT}.

\end{abstract}

\section{Introduction}
\label{sec:introduction}

Instruction tuning~\citep{ouyang2022training} is an important stage for large language models~(LLMs) after the knowledge-centric pre-training. It tailors pre-trained LLMs from a massive knowledge reservoir to a useful knowledge provider by training on instruction-following data~\cite{wang2022super}, endowing a significant boost on the LLMs' performance. Data synthesis techniques~\cite{wang2023self} play a key role in constructing such data. These techniques facilitate post-training in different domains due to their scalable nature, but meanwhile introduce more noises and duplications into the training data. \citet{zhou2024lima} find that models fine-tuned with only around 1K manually curated high-quality samples already demonstrate a strong generalization ability in downstream tasks. This finding encourages research into identifying the valuable subset for instruction tuning, which can lead to competitive or even superior performance while significantly reducing training costs.



\begin{figure}
    \centering
    \subfigure[\small Vanilla \label{fig:approach_vanilla}]{
    \begin{minipage}[t]{0.38\linewidth}
    \centering
    \includegraphics[width=\linewidth]{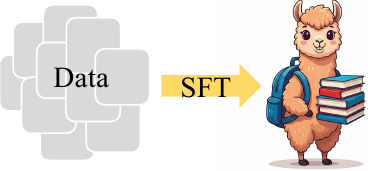}
    \end{minipage}%
    }%
    \subfigure[\small{Other Baselines}\label{fig:approach_baselines}]{
    \begin{minipage}[t]{0.6\linewidth}
    \centering
    \includegraphics[width=\linewidth]{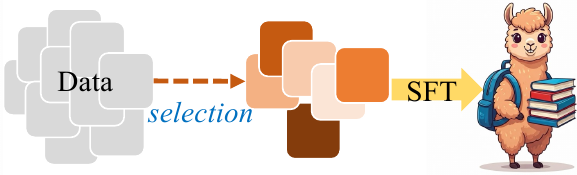}
    \end{minipage}%
    }%
    
    \subfigure[\small \ApproachName{} \label{fig:approach_our}]{
    \begin{minipage}[t]{1.0\linewidth}
    \centering
    \includegraphics[width=\linewidth]{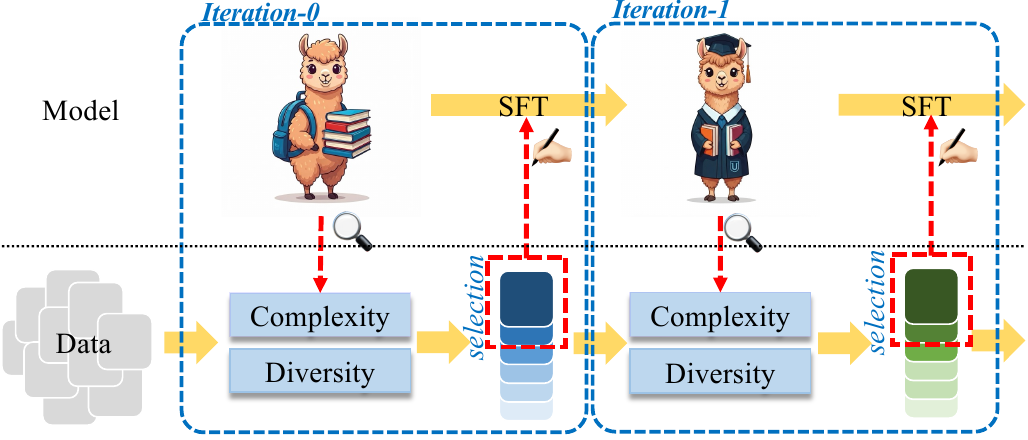}
    \end{minipage}%
    }%
    \caption{Illustrations of Vanilla, other baselines and \ApproachName{} . Grey boxes represent the training data that hasn't been assessed, which will be ranked by different metrics, i.e., the colored boxes. The red arrows in \ApproachName{} emphasize the collaboration between the model and the data. In other words, the model will supervise the data selection process, while the selected samples will be used to update the model's parameters.}
    \label{fig:approach}
\end{figure}

To automate the data selection process, certain studies~\cite{chen2023alpagasus,lu2023instag} employ powerful proprietary large language models, such as ChatGPT, to assess the quality of each sample based on predefined prompts. Nonetheless, these prompting methods incur additional costly expenses and lack interpretability. \citet{liumakes} developed scoring models using labels gathered from ChatGPT, and various approaches have been proposed to evaluate quality, either by leveraging a range of foundation models~\cite{liu2024selectit} or by focusing on a single model itself~\citep{li2024quantity,li-etal-2024-superfiltering,li-etal-2024-one}. 
Meanwhile, some research~\citep{shen2024rethinking,zhao2024long} also reveals that choosing samples by their response lengths is a simple but tough-to-beat baseline, even outperforming the elaborated metrics. 
With a deeper understanding of this task, researchers gradually reach a consensus on the importance of balancing the complexity and diversity of a selected subset~\citep{lu2023instag,liumakes}. Recent work by~\citet{wu2024best} has incorporated a complexity score that gauges the perplexity changes in the response, as derived from \citet{li-etal-2024-superfiltering}, and a TF-IDF-based diversity score computed on the instructions, leading to further enhancements.

However, we recognize that previous approaches have not fully harnessed the potential of both complexity and diversity metrics, nor have they truly integrated the strengths of both. On one hand, data is only selected at the outset of the fine-tuning process, which precludes model-specific scores from capturing the dynamic changes in the model. Based on our preliminary analysis using the instruction-following difficulty (IFD) score~\citep{li-etal-2024-superfiltering}, 55.31\% of the samples in the top 5\% of the selected subset after the first fine-tuning epoch were not included in the initial selection at the beginning of the fine-tuning process. On the other hand, previous studies~\citep{li2024quantity,wu2024best} suggest that the topic of a sample is predominantly influenced by its instruction. However, disparate instructions can result in similar and less informative responses. Since models are optimized using losses computed from response tokens, this limitation may impede the model's ability to function as an effective assistant.

To tackle the aforementioned shortcomings, we propose \ApproachName{}, which boosts LLMs' instruction-following performance through iterative data selection. An illustration with comparisons is in Fig.~\ref{fig:approach}.
Specifically, we employ the IFD metric to assess the complexity of each sample in a coarse-to-fine manner. Once the complexity scores are computed for all samples, we retain only the top-ranked ones for further selection, filtering out the remainder. 
Additionally, we compute the sum of TF-IDF features on each sample's response to determine the diversity score for each sample, aiming to identify the most diverse subset of informative responses. Once a sample is chosen, the diversity scores of the remaining samples are updated with weight decay applied to the features that have already been covered. Based on these two criteria, before each fine-tuning epoch, \ApproachName{} will update the complexity score of the reserved samples and greedily select a subset with the highest complexity-diversity score, continuing until the data of the specified budget is gathered. In extensive experiments conducted on instruction-tuning datasets of varying qualities, our \ApproachName{} demonstrated remarkably better performance, achieving results superior to those of a model trained with the full dataset using only 5\% of the data at each epoch, and consistently outperforming competitive baselines.

We summarize our contributions as follows:
\begin{itemize}
    \item We introduce \ApproachName{}, a new approach for instruction tuning by iterative data selection (Sec.~\ref{sec:approach}). \ApproachName{} shows favorable performance on tailoring a foundation language model to a useful assistant across different instruction-following datasets with extensive experiments (Sec.~\ref{sec:main-results}).

    \item Through experiments conducted on domain-specific datasets and other backbone LLMs, \ApproachName{} has shown strong generalization capabilities and consistently delivered improvements over baselines (Sec.~\ref{sec:domain-results}\&\ref{sec:other-model-results}).

    \item Through ablation studies and analysis of the selected data, we demonstrate that \ApproachName{} unleashes the power of complexity and diversity metrics, and underscores the critical importance of data-model collaboration during post-training (Sec.~\ref{sec:analysis}). 
\end{itemize}

    


\section{Approach}
\label{sec:approach}

The goal of supervised fine-tuning~(SFT) is to endow pre-trained foundation models with the ability to comprehend and follow user instructions.
Let $D$ denote the complete dataset containing $N$ samples $\{X_i, Y_i\}|_{i=0}^{N-1}$, where $X_i=\{x_{i,0}, ..., x_{i,A-1}\}$ refers to the instruction and $Y_i=\{y_{i,0}, ..., y_{i,B-1}\}$ refers to the answer. $A$ and $B$ denote the corresponding number of tokens. The process of supervised fine-tuning is to maximize the model's likelihood of generating answer $Y_i$ based on instruction $X_i$, i.e., $P_\theta(Y_i|X_i)$, where $\theta$ represents the LLM's parameters.

We propose \ApproachName{}, the core of which lies in a novel iterative data selection algorithm, with enhancements on measuring samples from both the complexity and diversity aspects. 
Our approach relies solely on the model being trained itself with affordable computation for data selection. An illustration is shown in Fig.~\ref{fig:approach_our}, with further details elaborated in the following sections.

    

\subsection{Complexity Measurement}

We characterize the complexity of a sample through instruction-following difficulty (IFD), as proposed by \citet{li-etal-2024-superfiltering}. This metric quantifies the connection between $X_i$ and $Y_i$ by assessing the perplexity of $Y_i$ using the model's prior knowledge directly, which is expressed as:

\begin{equation}
     {\rm PPL}_{\rm prior}^{i,\theta} = \exp{(-\frac{1}{B}\sum_{k=0}^{B-1} \log p_{\theta}(y_{i,k}|\boldsymbol{y}_{i,<k}))},
\end{equation}

and further conditioned on the corresponding instruction, as shown below:

\begin{equation}
     {\rm PPL}_{\rm cond}^{i,\theta} =\exp{(-\frac{1}{B}\sum_{k=0}^{B-1} \log p_{\theta}(y_{i,k}|\boldsymbol{y}_{i,<k}, X_i))}.
\end{equation}

The complexity score is then defined as the ratio of ${\rm PPL}_{\rm cond}^{i,\theta}$ to ${\rm PPL}_{\rm prior}^{i,\theta}$:

\begin{equation}
    \label{eq:complexity}
    S_{\rm COM}^{i,\theta} = \frac{{\rm PPL}_{\rm cond}^{i,\theta}}{{\rm PPL}_{\rm prior}^{i,\theta}}.
\end{equation}

This score serves to evaluate the effectiveness of instruction $X_i$ in facilitating the generation of $Y_i$. A lower value of $S_{\rm COM}^{i,\theta}$ suggests that the model can successfully follow the user's instruction without the need for additional training on that specific sample, while a higher value signifies greater complexity. Ideally, the following condition should be held:

\begin{equation}
    \forall i, \theta, \, {\rm PPL}_{\rm cond}^{i,\theta}< {\rm PPL}_{\rm prior}^{i,\theta},
\end{equation}

The rationale behind this is that the entropy of $Y_i$ should decrease when provided with more relevant information. If this condition is not met, the instruction-response pair is considered to be unaligned for the model.


\subsection{Diversity Measurement}

We base the diversity score of a selected subset on the informativeness of individual samples and the degree of information overlap among these samples. Specifically, the informativeness of a sample is predominantly manifested through its response, which conveys not only the topics of the sample but also the attitude in the response. A response rich in detailed information is generally more valuable and preferred by humans. Drawing inspiration from \citet{wu2024best}, we quantify the diversity score using the TF-IDF score of n-grams present in the response $Y_i$. Mathematically, the TF-IDF for a single n-gram $g$ is defined as:

\begin{equation}
    \begin{aligned}
    {\text{TF-IDF}}(g, Y_i) & = {\rm TF}(g, Y_i) \times {\rm IDF}(g, Y_i) \\
    & = \frac{f_g}{\sum_{k\in Y_i} f_k} \times \log\frac{N'}{N_g},
    \end{aligned}
\end{equation}
Here, $f$ denotes the count. $N'$ represents the size of the candidate set. 
$N_g$ refers to the number of samples in which the n-gram $g$ appears.

Subsequently, the diversity score of a sample $\{X_i, Y_i\}$ is:
\begin{equation}
\label{eq:informativeness}
    S_{\rm DIV}^i = \sum_{g\in Y_i} \alpha_g \times {\text{TF-IDF}}(g, Y_i).
\end{equation}
$\alpha$ is the weight newly introduced to measure the importance of the n-gram, which is initialized to 1. The higher the diversity score, the more informative the response is.

To select a diverse subset, we employ a greedy approach by dynamically adjusting the weight $\alpha$. Specifically, we reduce the weight of the n-grams in the selected samples at step $t$ by:
\begin{equation}
\label{eq:div-weight}
    \alpha_g^{t+1} = b \times \alpha_g^t,
\end{equation}
where $0\leq b<1$. In this way, samples covering diverse contents reflected by different lexical features will be selected with an increased possibility.

\subsection{Iterative Data Selection Algorithm}

Building upon the complexity and diversity scores defined earlier, we introduce an iterative data selection algorithm, dubbed \ApproachName{}, designed to effectively integrate both metrics for the purpose of identifying high-quality samples. Our approach adheres to the complexity-first and diversity-aware principle, which views the complexity—reflecting the pairwise relationship between the instruction and the response of a sample—as the foundational requirement. Simultaneously, it strives to maximize diversity among the candidates that already exhibit high complexity.

In alignment with the intuition to adapt to the evolving dynamics of the model during fine-tuning and to prevent the introduction of a substantial computational burden, we implement the complexity score in a coarse-to-fine, epoch-wise fashion. The complexity score for all samples is calculated at the beginning of the SFT process, and only the top-$(a\times M)$ samples are reserved for re-calculation and fine-grained selection after each epoch iteratively. $M$ refers to the number of samples to be selected for each epoch and $a>1$ is the coefficient for candidate re-calculation. Therefore, the time complexity for score calculation can be reduced from $O(\#{\rm steps}\times N)$ to $O(N+a\times M\times(\#{\rm epochs}-1))$, 
where both $\#{\rm epochs}\ll \#{\rm steps}$ and $M\ll N$.

Before each training epoch, we filter out the samples with $S_{\rm COM}^{i,\theta_t}\geq 1$ among the $a\times M$ candidates, and calculate the comprehensive score for each sample as follows:
\begin{equation}
    S^i = S_{\rm COM}^{i,\theta_t} \times S_{\rm DIV}^{i}.
\end{equation}
The sample with the highest $S^i$ will be selected. Subsequently, we update the $S_{\rm Div}^{i}$ according to Eq.~\ref{eq:div-weight}, which will in turn impact $S^i$. \ApproachName{} repeats this process greedily, until a total of 
M samples are collected for the instruction-tuning phase of the current epoch.

Overall, our data selection algorithm, designed to balance the dual objectives of effectiveness and efficiency in instruction tuning, is achieved in an iterative manner. The algorithm is elaborated in Algorithm~\ref{alg:IDS}.


\begin{algorithm}[t]
	\caption{The iterative data selection algorithm.}
	\label{alg:IDS}
	\textbf{Input}: the pre-trained model $\mathcal{M}_{\theta_0}$, the initial instruction tuning dataset $D$\\
	\textbf{Parameter}: the number of epochs $t\in\{0, ..., T-1\}$, the number of samples selected in each epoch $M$, the re-calculation parameter $a$, the weight decay parameter $b$.\\
	\textbf{Output}: the instruction-tuned model $\mathcal{M}_{\theta_T}$

	\begin{algorithmic}[1] 

			
		\FOR{training epoch $t=0,..., T-1$}
			\STATE \textit{// Data selection process}
                \STATE Calculate $S_{\rm COM}^{i,\theta_t}$ for all samples in $D$ 
                \IF{t==0}
                \STATE Sort samples in $D$ based on $S_{\rm COM}^{i,\theta_t}$
                \STATE $D = D[:a\times M]$
                \ENDIF

                \STATE $D' = \{(X_i, Y_i)\mid (X_i, Y_i)\in D, S_{\rm COM}^{i,\theta_t}<1 \}$
                \STATE Instruction-tuning data for the current epoch $D_t = \emptyset$
                \WHILE{$|D_t|<M$}
                    \STATE Calculate $S_{\rm DIV}^i$ for all samples in $D'$
                    \STATE Calculate the comprehensive score $S^i$ for all samples in $D'$
                    \STATE $D_t = D_t  \cup \{ (X_i, Y_i) \in D' : S^i = \max_{(X_j, Y_j) \in D'} S^j \}$
                    \STATE Update $S_{\rm DIV}^i$ according to Eq.~\ref{eq:div-weight}
                \ENDWHILE
                
                \STATE \textit{// Training process}
                
                \FOR{training steps in an epoch}
				\STATE Randomly sample a batch $B$ from $D_t$
                    \STATE Update $M_{\theta_t}$ by the Cross Entropy Loss.	
                \ENDFOR	
            \ENDFOR
		\STATE \textbf{return} $M_{\theta_{T-1}}$
	
	\end{algorithmic}

\end{algorithm}
\section{Experimental Setup}
\label{sec:experiment}

\subsection{Instruction-tuning Datasets}

We conduct instruction tuning upon LLMs with four different instruction-tuning datasets. \textbf{Alpaca}~\citep{alpaca} contains 52,000 samples that are created by leveraging text-davinci-003 model under the self-instruct framework~\citep{wang2023self}. \textbf{Alpaca-GPT4}~\citep{peng2023instruction} contains higher quality responses generated by GPT-4 given the same instructions from Alpaca.
\textbf{WizardLM}~\citep{xu2023wizardlm}  refers to the 70K evolved samples collected by the Evol-Instruct algorithm that rewrites the initial instruction from Alpaca step by step into more complex instruction by ChatGPT. \textbf{Dolly}~\citep{conover2023free} consists of 15K human-generated prompt-response pairs for instruction tuning.

\subsection{Evaluation Benchmarks}

To evaluate the capabilities of instruction-tuned LLMs comprehensively, we selected widely-adopted benchmarks across a spectrum of targeted abilities. They include GSM8K~\citep{cobbe2021training} for arithmetic reasoning, MMLU~\citep{hendrycksmeasuring} for factual knowledge, TruthfulQA~\citep{lin2022truthfulqa} for safety, BBH~\citep{suzgun2023challenging} for multi-step reasoning and HumanEval~\citep{chen2021evaluating} for coding capability, ARC~\citep{clark2018think} for scientific questions and Hellaswag~\citep{zellers2019hellaswag} for commonsense understanding. To guarantee the fairness of evaluation, all of the models are evaluated using publicly available code bases, including open-instruct~\footnote{\url{https://github.com/allenai/open-instruct}} and Open LLM Leaderboard~\footnote{\url{https://github.com/EleutherAI/lm-evaluation-harness/}}.

Moreover, we employ MixEval~\citep{ni2024mixeval}, which adeptly captures the breadth and subtlety of real-world user queries, demonstrating a 0.96 correlation with Chatbot Arena. Specifically, we conducted our evaluation using MixEval-hard-0601 to gauge the model's proficiency in handling general user queries.



\subsection{Baselines}

We compared our approach with five representative baselines as follows:
\begin{itemize}[leftmargin=0pt, nolistsep, itemindent=2em, label=$\circ$]
    \item \textbf{Vanilla} refers to supervised fine-tuning with the whole dataset. 
    \item \textbf{Longest}~\cite{zhao2024long,shen2024rethinking} is a rule-based method that selects the samples with the longest response.
    \item \textbf{Deita}~\cite{liumakes} trains scoring models based on labels collected from ChatGPT for complexity and quality assessments, and measures diversity via distances of model embeddings. The samples are selected greedily with the highest complexity-quality scores while not being redundant with the others.
    \item \textbf{Superfiltering}~\cite{li-etal-2024-superfiltering} improves the model-specific IFD score~\cite{li2024quantity} and proposes to select samples with the highest IFD score.
    \item \textbf{GraphFilter}~\cite{wu2024best} utilize the IFD score for complexity and TF-IDF scores based on instructions for diversity. The data are selected greedily by the priority score defined as the multiplication of both metrics.
\end{itemize}

For fair comparisons, we calculate the IFD scores using the target model itself instead of a smaller model such as GPT-2~\cite{radford2019language}. Besides, the same number of samples are selected by different approaches and all of the models are updated for the same steps during fine-tuning.

\begin{table*}[h]
    \centering
    \caption{Performance of approaches backed on Llama-3-8B fine-tuned with general instruction-following datasets. The highest score in each column is in bold, and the second-best ones for overall performance are underlined. $\star$ marks our proposed approach. All of the experiments are re-implemented upon the same pre-trained model for fair comparisons.}
    \scalebox{1.0}{\begin{tabular}{l|ccccccc|c|c}
        \toprule[1pt]
         Method & GSM8K & MMLU & \makecell{Truthful\\QA} & BBH & \makecell{Human\\Eval} & ARC & \makecell{Hella\\Swag} & AVG & MixEval\\
        \midrule[1pt]
        \multicolumn{10}{l}{\textit{Results on Alpaca}} \\
        Vanilla & 20.00 & 52.20 & 42.78 & 48.52 & 28.05 & 58.53 & 81.77 & 47.41 & 27.80 \\
        Longest & 56.50 & 59.12 & 43.12 & 57.59 & 42.80 & 62.20 & \textbf{83.58} & \underline{57.84} & \textbf{35.60}\\
        Deita & 37.00 & 56.11 & 42.20 & 52.69 & 38.66 & 64.76 & 83.00 & 53.49 & 29.05 \\
        Superfiltering & 51.50 & 52.09 & \textbf{43.72} & 55.19 & 38.54 & 62.03 & 83.37 & 55.21 & 31.20 \\
        GraphFilter & 44.50 & \textbf{59.59} & 43.57 & 57.31 & 42.32 & \textbf{65.19} & 83.12 & 56.51 & 29.35 \\
        \ApproachName{}$\star$ & \textbf{61.00} & 59.31 & 40.81 & \textbf{59.44} & \textbf{46.10} & 61.43 & 83.11 & \textbf{58.74} & \textbf{35.60} \\
        \midrule[1pt]
        \multicolumn{10}{l}{\textit{Results on Alpaca-GPT4}} \\
        Vanilla & 41.00 & 50.78 & 52.69 & 48.61 & 39.76 & 59.98 & 82.42 & 53.61 & 34.15 \\
        Longest & 60.60 & 59.93 & 54.61 & 56.67 & 42.20 & 61.69 & \textbf{84.18} & 59.97 & \textbf{40.60} \\
        Deita & 56.50 & 56.39 & 52.56 & 51.48 & 37.80 & 63.14 & 84.09 & 57.42 & 33.10 \\
        Superfiltering & 59.00 & \textbf{60.55} & 56.19 & 57.04 & 44.27 & 62.80 & 83.84 & 60.53 & 36.35 \\
        GraphFilter & 62.50 & 58.62 & \textbf{58.58} & 57.22 & \textbf{46.22} & \textbf{63.23} & 83.81 & \underline{61.45} & 33.00\\
        \ApproachName{}$\star$ & \textbf{68.50} & 60.35 & 56.93 & \textbf{60.37} & 44.63 & 60.67 & 83.92 & \textbf{62.20} & \underline{40.15} \\
        \midrule[1pt]
        \multicolumn{10}{l}{\textit{Results on WizardLM}} \\
        Vanilla & 56.00 & 53.15 & 49.68 & 53.43 & 45.12 & 57.17 & 81.74 & 56.61 & 35.30 \\
        Longest & 60.00 & 59.88 & 48.43 & 57.78 & \textbf{49.39} & 59.81 & 83.09 & 59.75 & \underline{36.55} \\
        Deita & 62.00 & 56.68 & 47.28 & 54.17 & 48.45 & 61.43 & 83.19 & 59.03 & 31.10 \\
        Superfiltering & 56.50 & 59.17 & \textbf{50.97} & 59.81 & 47.68 & 61.26 & 83.00 & \underline{59.77} & 34.50 \\
        GraphFilter & 58.50 & \textbf{60.79} & 47.06 & \textbf{60.56} & 46.22 & 61.35 & 83.02 & 59.64 & 33.00 \\
        \ApproachName{}$\star$ & \textbf{62.00} & 60.33 & 50.73 & 58.80 & 45.24 & \textbf{61.69} & \textbf{83.27} & \textbf{60.29} & \textbf{37.20} \\
        \bottomrule[1pt]
        
    \end{tabular}}
    \label{tab:main-results}
\end{table*}

\subsection{Implementation Details}

We mainly carried out experiments on the LLaMA-3-8B pre-trained language model~\cite{dubey2024llama}. All of the models are trained with batch size equaling 32 for 3 epochs. The learning rate is set to 2e-5 following the training configuration in the Alpaca codebase~\footnote{\url{https://github.com/tatsu-lab/stanford_alpaca}}. \ApproachName{} selects only 5\% of the whole training set for fine-tuning, i.e., $M=0.05 * N$. We set $a=3$ for fine-grained candidate preparation. $b$ equals 0.1 for Alpaca, WizardLM and Dolly, and 0.0 for Alpaca-GPT-4. Other setup variations are specified in the following sections.

\section{Results}
\label{sec:results}

We first present the overall comparisons with competitive baselines on different instruction-following datasets, followed by generalization performance on domain-specific datasets and other backbone models.

\subsection{Performance on General Instruction-following Datasets}
\label{sec:main-results}

We compared models trained with different data selection approaches on various datasets. The results on Dolly are in Appendix~\ref{app:dolly-results} due to space limitation.


\textbf{Results on individual benchmarks}\quad As shown in Table~\ref{tab:main-results}, none of the approaches consistently outperform the others among different benchmarks. Superior scores on one or two datasets also don't necessarily lead to the best overall performance. For instance, GraphFilter achieved leading performance with 59.59\% and 65.19\% accuracy on MMLU and ARC, respectively. Nevertheless, its overall performance, as reflected by the average score and MixEval, falls outside the top-2 approaches. In other words, data selection methods can easily achieve significant improvement in a single task when the selected data is closer to the distribution of a specific group of test data. Therefore, the overall performance of the model is more important for evaluating the effectiveness of the method, and single-task performance can be regarded as an indicator of detailed abilities.

\textbf{Results on overall performance}\quad According to the average scores across the seven benchmarks and the MixEval benchmark, we have the following observations:

\begin{table*}[t!]
    \centering
    \caption{Performance of approaches backed on Llama-3-8B fine-tuned with CodeAlpaca.}
    \begin{tabular}{l|ccc|ccc|c}
        \toprule[1pt]
        \multirow{2}{*}{Method} & \multicolumn{3}{c}{HumanEval} & \multicolumn{3}{c}{MBPP} & \multirow{2}{*}{AVG} \\
         & pass@1 & pass@5 & pass@10 & pass@1 & pass@5 & pass@10 & \\
         \midrule[1pt]
        Vanilla & 38.35 & 44.41 & 45.73 & 39.71 & 40.85 & 41.01 & 41.68 \\
        Longest & \textbf{42.87} & 49.82 & 52.44 & 36.98 & 44.27 & 46.06 & \underline{45.40} \\
        Deita & 35.37 & 41.79 & 44.51 & 38.45 & 42.67 & 43.17 & 40.99 \\
        Superfiltering & 36.77 & 44.86 & 48.17 & 36.62 & 40.61 & 41.73 & 41.46 \\
        GraphFilter & 42.26 & 48.04 & 50.00 & 39.71 & 43.37 & 44.24 & 44.60 \\
        \ApproachName{}$\star$ & 41.04 & \textbf{49.83} & \textbf{54.27} & \textbf{40.65} & \textbf{46.56} & \textbf{48.20} & \textbf{46.76} \\
        \bottomrule[1pt]
    \end{tabular}
    \label{tab:code-results}
\end{table*}

Previous data selection methods struggle to achieve consistent improvements over Vanilla among different datasets. Deita outperforms Vanilla by 1.25\% on MixEval trained on the Alpaca dataset, while lags behind Vanilla by 1.05\% and 4.2\% on Alpaca-GPT4 and WizardLM correspondingly, where the instruction-response pairs are of better quality. The same trends are reflected by Superfiltering and GraphFilter.

Longest shows superior performance compared to other baselines. Although it does not perform exceptionally well on most of the single-dimension benchmarks, it shows favorable performance in the overall metrics. Longest not only consistently outperforms Vanilla with a mere 5\% of original data, but also surprisingly beats the other data selection methods that require additional computations for selecting a high-quality subset.

Our proposed \ApproachName{} demonstrates consistent improvements over Vanilla on various instruction-tuning datasets and beats the strongest rule-based approach, Longest, on most of the metrics.  It also shows stronger generalization ability than Longest, as discussed in the following sections. On the other hand, although complexity and diversity have been explored in other baselines, \ApproachName{} further enhances the benefits of combining complexity and diversity by emphasizing the synergy between the model and the data, thereby achieving state-of-the-art performance.


\subsection{Performance on Domain-Specific Data}
\label{sec:domain-results}

We further investigate the effectiveness of various data selection approaches on domain-specific datasets. Specifically, we select CodeAlpaca~\citep{chaudhary2023code}, an instruction tuning dataset collected in a manner similar to Alpaca, which is designed to improve the code generation capabilities of LLMs. We utilize the HumanEval~\cite{chen2021evaluating} and MBPP+~\cite{evalplus} benchmarks, which are evaluated using the pass@$k$ metrics, indicating the success rate of a model within $k$ attempts.

According to the results presented in Table~\ref{tab:code-results}, \ApproachName{} exhibits superior performance in code generation scenarios. It consistently outperforms Longest, with the exception of pass@1 on HumanEval. Furthermore, among all approaches, \ApproachName{} and GraphFilter are the only two that consistently surpass Vanilla across all evaluation metrics.

\subsection{Performance on Other Backbone Model}
\label{sec:other-model-results}

To assess the generalization capability across different backbone pre-trained models, we conducted additional experiments on Qwen-2.5-7B~\citep{qwen25}, and compared \ApproachName{} with Vanilla and the strongest baseline, Longest. The overall results are listed in Table~\ref{tab:qwen-results}, with more details elucidated in Appendix~\ref{app:qwen-results}. \ApproachName{} consistently outperforms both Longest on most metrics and Vanilla.

\begin{table}[h!]
    \centering
    \caption{Performance of Approaches backed on Qwen-2.5-7B fine-tuned with general instruction-following datasets.}
    \label{tab:qwen-results}
    \begin{tabular}{p{1.0cm}c|ccc}
        \toprule[1pt]
         Dataset & Benchmark & Vanilla & Longest & \ApproachName{} \\
         \midrule[1pt]
         \multirow{2}{*}{Alpaca} & AVG & 65.17 & \textbf{66.49} & 66.43 \\
         & MixEval & 38.30 & 38.45 & \textbf{38.75} \\
         \hline
         \multirow{2}{*}{\makecell{Alpaca\\-GPT4}} & AVG & 68.96 & 68.66 & \textbf{69.04} \\
         & MixEval & 40.15 & 39.50 & \textbf{40.95} \\
         \hline
         \multirow{2}{*}{\makecell{Wizard\\LM}} & AVG & 69.17 & 69.05 & \textbf{69.32}\\
         & MixEval & 38.20 & 39.50 & \textbf{40.50}\\
         
         \bottomrule[1pt]
    \end{tabular}
\end{table}


\section{Ablations and Analysis}
\label{sec:analysis}
In this section, we analyze the rationale behind the design of \ApproachName{} with ablation studies and data visualization. First, we show the effectiveness of iterative selection. Following that, we conduct ablations to establish the significance of incorporating diversity in our complexity-first data selection algorithm. Next, we analyze the characteristics of the selected data. Lastly, we assess the sensitivity of hyper-parameters newly introduced in \ApproachName{}.

\subsection{Ablation for Iterative Selection}

\begin{figure}[t]
    \centering
    \includegraphics[width=0.9\linewidth]{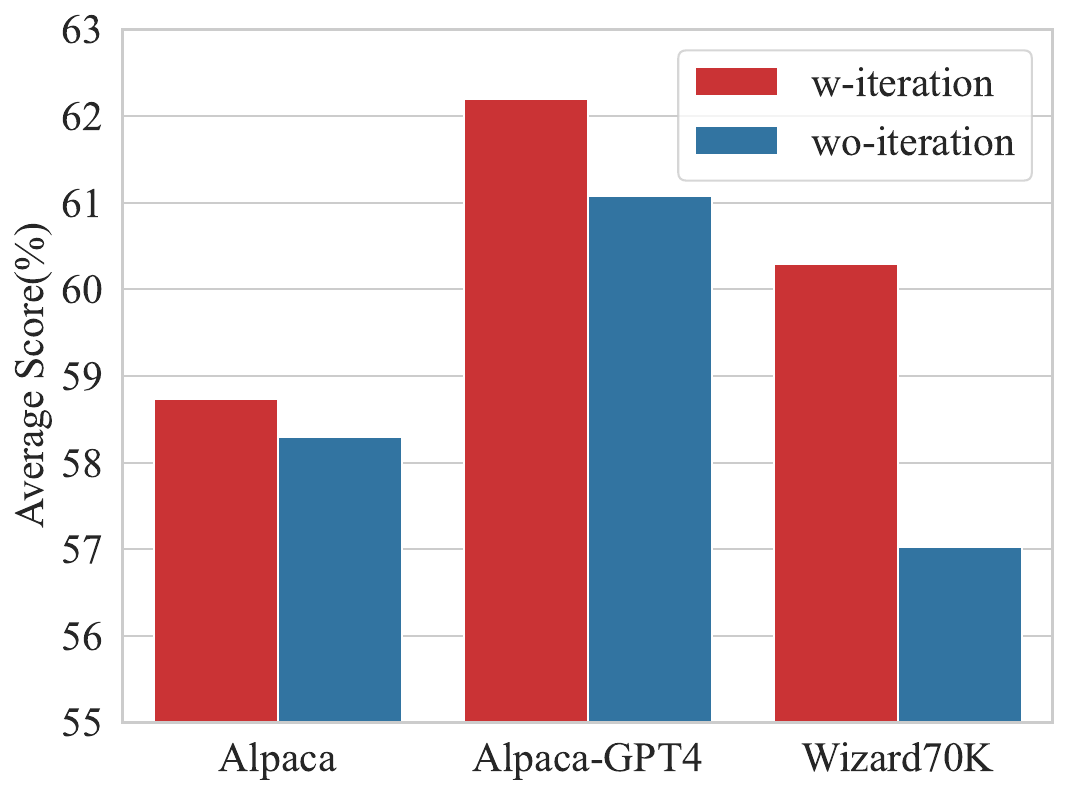}
    \caption{Ablation on the need of iterative selection. Models are evaluated by the average performance(\%) over 7 datasets.}
    \label{fig:iteration-ablation}
\end{figure}

We verify the necessity of performing iterative selection by comparing the performance of our complete approach (\textbf{w-iteration}) with the ablation (\textbf{wo-iteration}), which uses a fixed subset selected at the beginning of the fine-tuning process. Figure~\ref{fig:iteration-ablation} indicates that the average performance over 7 benchmarks consistently drops when removing the iterative selection operation, demonstrating the importance of catering to the dynamic changes of the model by updating model-specific complexity scores during the fine-tuning process.

\subsection{Ablation for Introducing Diversity}

To analyze the importance of incorporating the diversity measurement into the complexity-first iterative data selection algorithm, we conduct experiments on the Alpaca dataset with the following ablations:
\begin{itemize}[leftmargin=0pt, nolistsep, itemindent=2em, label=$\circ$]
    \item \textbf{``w/o div''} refers to the algorithm that does not incorporate $S_{\rm DIV}^i$ and selects samples based on $S_{\rm COM}^{i,\theta}$.
    \item \textbf{``w div($\cdot$)''} refers the algorithm that utilizes $S_{\rm DIV}^i$ calculated on different part of the data. $i$, $o$, $i+o$ represents $X_i$, $Y_i$ and the concatenation of $(X_i, Y_i)$, respectively.
\end{itemize}

Results are listed in Table~\ref{tab:ablation-diversity}. ``w/o div'' lags behind most of the other ablations considering the diversity metric, indicating the significance of making a balance between complexity and diversity. Using $S_{\rm DIV}^i$ based solely on the instruction does not help enhance the model's overall performance, as evidenced by the lowest average score. Our approach \ApproachName{}, i.e., ``w div($o$)'', shows superior performances among ablations, highlighting the importance of defining the diversity of a sample based on its response. It achieves a notable improvement on GSM8K, MMLU, BBH and HumanEval compared to ``w/o div''. 

\begin{table*}[h!]
    \centering
    \caption{Ablation performance for the incorporation of diversity metric. The best results are in \textbf{bold}.}
    \begin{tabular}{l|ccccccc|c}
        \toprule[1pt]
        Ablations & GSM8K & MMLU & TruthfulQA & BBH & HumanEval & ARC & HellaSwag & AVG \\
        \midrule[1pt]
        w/o div & 57.5 & 53.90 & 43.01 & 57.31 & 36.83 & \textbf{62.46} & 83.02 & 56.29 \\
        w div($i$) & 46.00 & 57.53 & \textbf{45.83} & 56.30 & 41.10 & 61.35 & 82.72 & 55.83 \\
        w div($i+o$) & 56.50 & 58.32 & 42.43 & 57.87 & 41.46 & 61.95 & 82.39 & 57.27\\
        w div($o$) & \textbf{61.0} & \textbf{59.31} & 40.81 & \textbf{59.44} & \textbf{46.10} & 61.43 & \textbf{83.11} & \textbf{59.74} \\
        \bottomrule[1pt]
    \end{tabular}
    \label{tab:ablation-diversity}
\end{table*}

\subsection{Analysis on Selected Data}
\label{sec:analysis-selected-data}

We dive deeper into analyzing the characteristics of the selected instruction tuning data by different approaches. Considering the strong performance of Longest, we first analyze the response length of different approaches measured by the number of words, as well as the three subsets selected by our \ApproachName{} for each epoch. The statistics for Alpaca are shown in Fig.~\ref{fig:length-analysis}, with additional results in Appendix~\ref{app:response-length}.

The overall trend among different approaches is Longest$>$\ApproachName{}$>$Superfiltering$>$GraphFilter$>$Vanilla$>$ Deita, similar to the performance trend show in Table~\ref{tab:main-results}. Nevertheless, the reversed pairwise relations (Longest, \ApproachName{}) and (Deita, Vanilla) show that length is not the only golden criterion for data selection. Our approach selects data based on complexity and diversity,  outperforming Longest with smaller response lengths.


\begin{figure}[t]
    \centering
    \includegraphics[width=\linewidth]{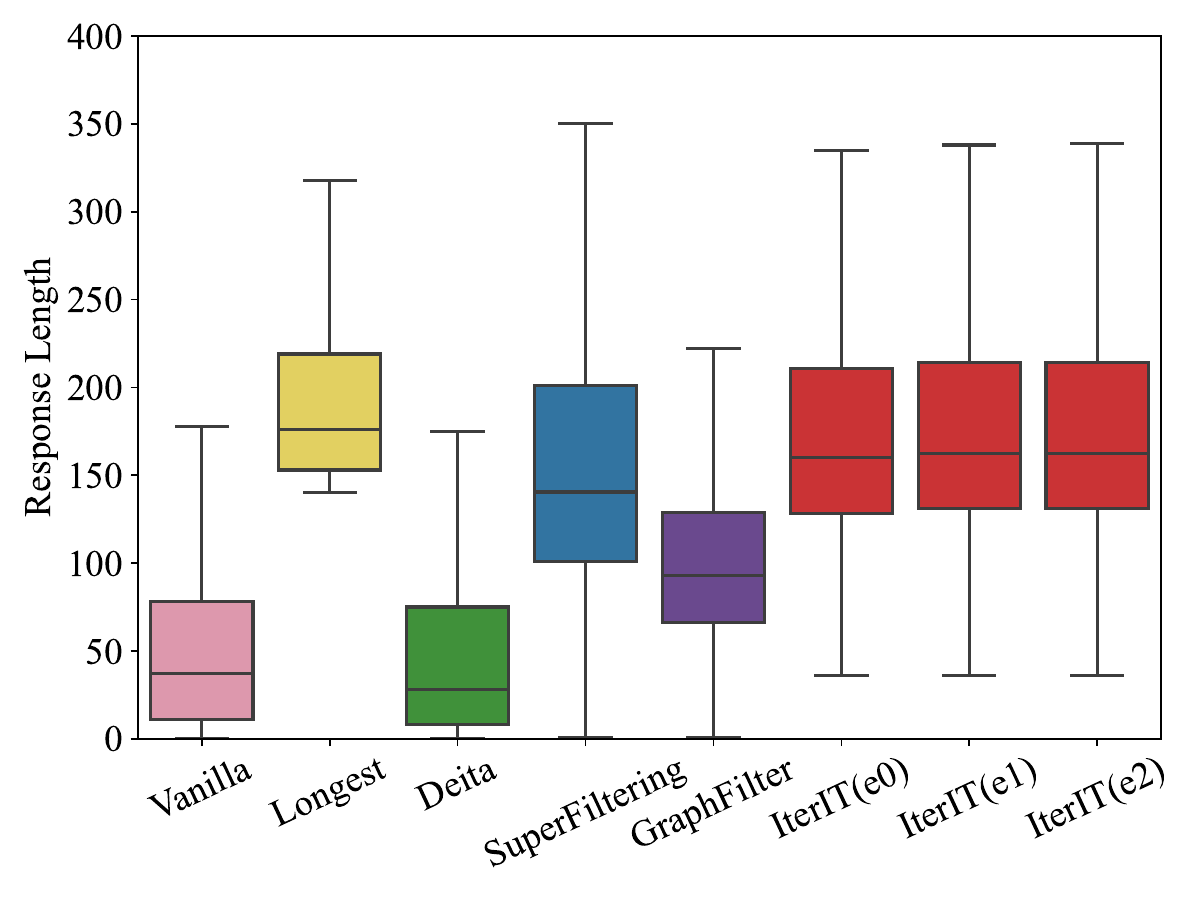}

    
    
    \caption{The response length of instruction-tuning data selected from Alpaca by different approaches.}
    \label{fig:length-analysis}
\end{figure}

We further analyze the similarity between the selected subsets by different approaches. The Jaccard similarity between the subset of Longest and the ones used in each epoch by \ApproachName{} is all around 50\%. Besides, the lower quartile of \ApproachName{} is much lower than that of Longest in Fig.~\ref{app:response-length}. These observations suggest that samples with shorter responses can also be valuable for instruction tuning, which is not only a key difference between the selected data but also contributes to the favorable performance of \ApproachName{}.
In addition, the similarity between the subsets in our approach is listed in Table~\ref{tab:jaccard-sim}. The Jaccard similarity is much lower between Epoch 1 and Epoch 3, showing the model's preference to the instruction-tuning data changes as more update steps are carried out. \ApproachName{} made adjustments to the data selection during fine-tuning for catering to the dynamics of the model. 

\begin{table}[]
    \centering
    \caption{Jaccard similarity(\%) between pairs of subset selected by \ApproachName{} during instruction tuning.}
    \begin{tabular}{c|ccc}
    \toprule[1pt]
         Epochs& 1,2 & 2,3 & 1,3  \\
    \midrule[1pt]
         Jaccard Similarity & 78.88 & 79.43 & 70.55 \\
    \bottomrule[1pt]
    \end{tabular}
    \label{tab:jaccard-sim}
\end{table}




\subsection{Sensitivity of Hyper-parameters}

We discuss the effects of hyper-parameters in \ApproachName{}, including the selected data size for each epoch, the candidate size for re-calculation controlled by $a$, and the weight decay coefficient $b$ for diversity measurement. 

\begin{figure*}[h!]
    \centering
    \subfigure[\small Selection Size \label{fig:analysis_data_size}]{
    \begin{minipage}[t]{0.3\linewidth}
    \centering
    \includegraphics[width=\linewidth]{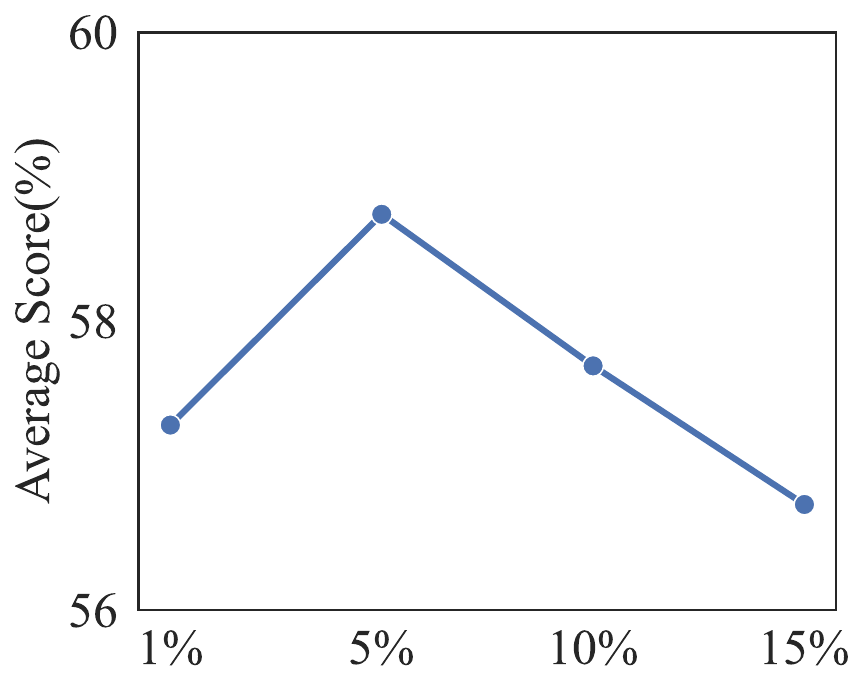}
    \end{minipage}%
    }%
    \subfigure[\small{$a$ for Candidate Size}\label{fig:analysis_candidate_size}]{
    \begin{minipage}[t]{0.3\linewidth}
    \centering
    \includegraphics[width=\linewidth]{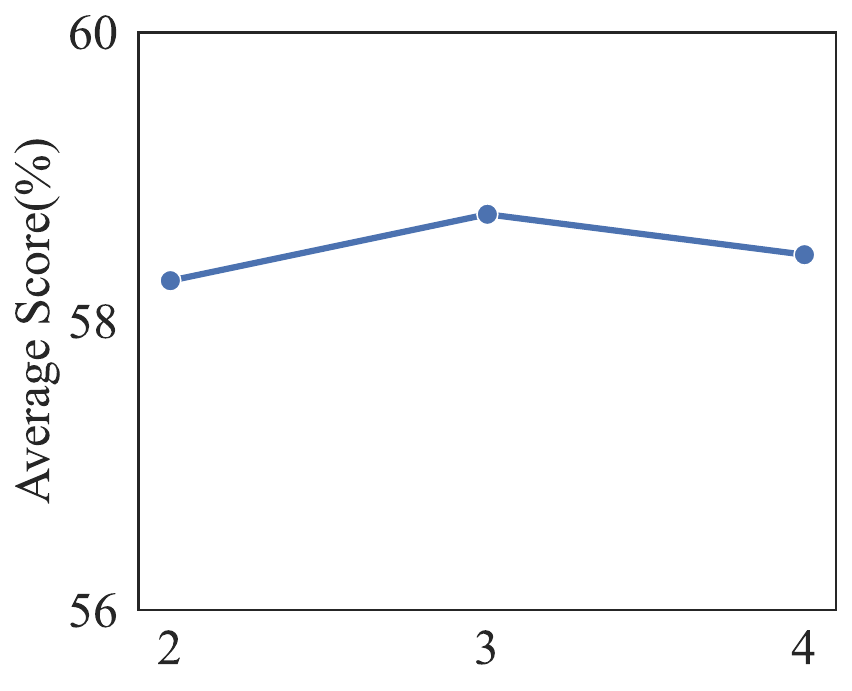}
    \end{minipage}%
    }%
    \subfigure[\small Weight Decay $b$\label{fig:analysis_weight_decay}]{
    \begin{minipage}[t]{0.3\linewidth}
    \centering
    \includegraphics[width=\linewidth]{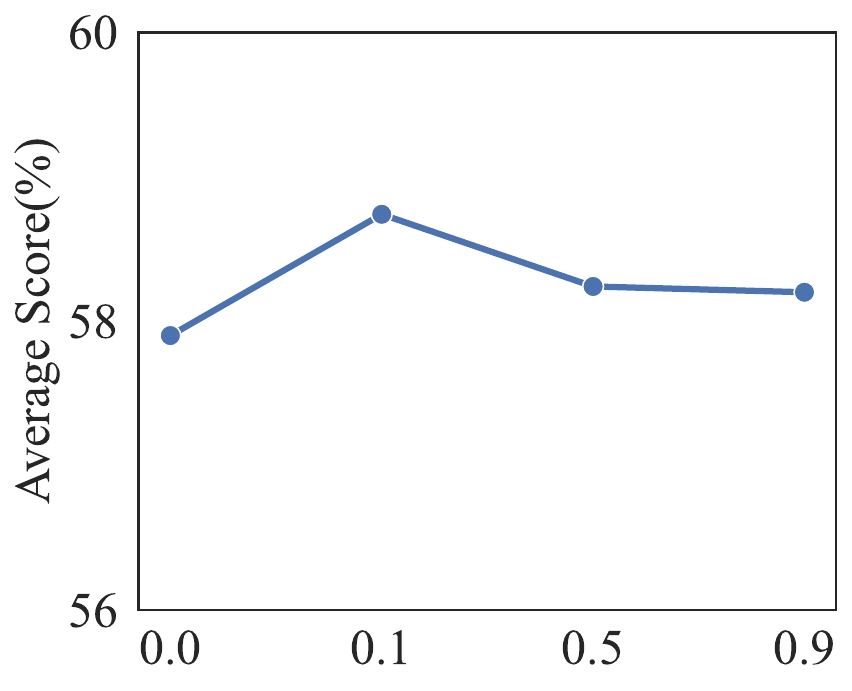}
    \end{minipage}%
    }%
    \caption{The average scores(\%) of models trained on Alpaca under different hyper-parameters.}
    \label{fig:length-analysis-2}
\end{figure*}

\textbf{Data Size} The model's performance with selection size $M$ set to different percentages of the original dataset is illustrated in Fig.~\ref{fig:analysis_data_size}.
The performance does not exhibit a positive correlation with the amount of selected data. When only 1\% of the data is selected, \ApproachName{} requires a greater number of training epochs to achieve convergence, resulting in frequent recalculation processes and a reduction in training efficiency. Meanwhile, the performance does not improve even when the model is trained on more data with additional steps. At this point, the recalculation process cannot be executed in a timely manner, which hinders the synergy between the model and the data, leading to suboptimal performance. Therefore, we suggest setting $M$ in the range of 1K to 5K considering the balance between training efficiency and the model's performance. For fair comparisons, we set $M=0.05N$ across different datasets in this paper.

\textbf{Candidate Pool} Performance under different candidate sizes calculated by the multiplication of $a$ and $M$ for recalculation, is shown in Fig.~\ref{fig:analysis_candidate_size}. Enlarging the candidate size not only lifts the computational load, but also increases the likelihood of low-quality data being selected due to the imperfect complexity or diversity score. Therefore, we propose a coarse-to-fine way for updating the complexity score and suggest setting $a=3$. In this way, low-quality samples recognized at the beginning will be removed directly, and sufficient high-quality samples will be considered for diversity consideration.

\textbf{Weight Decay} The rationale behind introducing weight decay in the diversity score $S_{\rm DIV}^i$ is illustrated in Fig.~\ref{fig:analysis_weight_decay}. We hypothesize that removing the TF-IDF scores for already selected n-grams may not be appropriate for mathematical data and code data. These samples contain reserved words that have a high repetition rate. Setting $b$ to 0 reduces the likelihood of such data being selected, which may adversely affect the reasoning capabilities of LLMs. This can be further verified by the 2.50\%, 2.59\% and 4.76\% improvement of $b=0.1$ on GSM8K, BBH and HumanEval, respectively, compared with $b=0.0$.
On the other hand, adopting a large $b$ will have a negative impact on the diversity of data. Therefore, we set $b=0.1$ for most instruction-tuning scenarios involving data from various tasks, and suggest $b=0.0$ for task-specific scenarios where reserved words are either undefined or shared across all samples.

\section{Related Work}
\label{sec:related_work}

\subsection{Instruction Tuning Datasets}

Instruction tuning teaches LLMs to perceive the intent of users and provide helpful responses, standing as a core component in the deployment of LLMs~\cite{ouyang2022training}. Collecting high-quality data for instruction tuning has caught great attention. Early works~\cite{weifinetuned,longpre2023flan} merge existing NLP datasets to obtain a diverse collection across different tasks. Subsequently, \citet{wang2023self} proposed the Self-Instruct framework, which is an automated algorithm gathering instruction-response data by bootstrapping LLMs' generations. Building on this work, a number of data synthesis and evolution techniques~\cite{xu2023wizardlm,ding2023enhancing,wu2024lamini,li2024selective} have emerged, leveraging powerful proprietary LLMs to significantly enhance the quality and size of candidate datasets for supervised alignment. Nonetheless, according to the superficial alignment hypothesis proposed by \citet{zhou2024lima}, LLMs have acquired abundant knowledge and abilities during pre-training, and the focus of instruction tuning is all about style learning for providing a helpful response. They verified their hypothesis with around 1K elaborately selected samples, highlighting data selection for instruction tuning as a promising research direction.


\subsection{Data Selection for Instruction Tuning}
Although it's widely accepted that the quality of instruction tuning data is more significant than the quantity, the criteria for quality measurement remains mysterious. AlpaGasus~\cite{chen2023alpagasus} relies on ChatGPT's understanding and prompts it to score samples based on its quality. \citet{cao2023instruction} relies on a bag of indicators measuring quality from different aspects, such as length, coherence, understandability, etc., trying to estimate the models' performance trained on a selected subset. Nuggests~\cite{li-etal-2024-one} leverages one-shot learning performance to discern the quality of the data, and SelectIT~\cite{liu2024selectit} utilizes the intrinsic uncertainty of LLMs from different levels, including token, sentence and model, to make a collaborative decision. Meanwhile, research from \citet{shen2024rethinking} and \citet{zhao2024long} argue that selecting the longest responses is a simple but tough-to-best baseline, which aligns with the superficial alignment hypothesis. They find that the longest samples are more challenging, high-quality and preferred by humans.

More works consider complexity and diversity for assessing data quality. InsTag~\cite{lu2023instag} queries proprietary models to tag samples based on semantics and intentions, and defines diversity and complexity measurements regarding the number of tags. \citet{li-etal-2024-superfiltering} measures the complexity from the aspect of instruction-following difficulty and proposes the model-specific IFD score~\cite{li2024quantity}. Deita~\cite{liumakes} trains an additional scoring model based on labels collected from ChatGPT from the aspects of complexity and quality, and considers diversity via distance between samples' semantic representations during the selection process. \citet{wu2024best} considers lexical diversity based on bipartite graph and IFD score at the same time, outperforms a number of baselines~\cite{wang2024interpretable,arthur2006k,li-etal-2024-superfiltering} that mainly consider a single dimension. Our approach \ApproachName{} is in line with these works and aims at advancing instruction tuning performance in an efficient way.






\section{Conclusion}
\label{sec:conclusion}

We introduce \ApproachName{} for improving LLMs' instruction tuning by iterative data selection. Our approach successfully unleash the power of quality metrics by perceiving the model dynamics for complexity calculation and exploiting response informativeness for diversity calculation. By integrating both aspects, our approach demonstrates superior performance on multiple instruction tuning datasets and generalizes well to domain-specific scenarios. We underscore the importance of model-data collaboration towards powerful LLMs and will devote into more complicated scenarios and different training stages of LLMs in the future.

\bibliography{example_papers}

\begin{thebibliography}{41}
\providecommand{\natexlab}[1]{#1}
\providecommand{\url}[1]{\texttt{#1}}
\expandafter\ifx\csname urlstyle\endcsname\relax
  \providecommand{\doi}[1]{doi: #1}\else
  \providecommand{\doi}{doi: \begingroup \urlstyle{rm}\Url}\fi

\bibitem[Arthur \& Vassilvitskii(2006)Arthur and Vassilvitskii]{arthur2006k}
Arthur, D. and Vassilvitskii, S.
\newblock k-means++: The advantages of careful seeding.
\newblock Technical report, Stanford, 2006.

\bibitem[Cao et~al.(2023)Cao, Kang, Wang, and Sun]{cao2023instruction}
Cao, Y., Kang, Y., Wang, C., and Sun, L.
\newblock Instruction mining: Instruction data selection for tuning large
  language models.
\newblock \emph{arXiv preprint arXiv:2307.06290}, 2023.

\bibitem[Chaudhary(2023)]{chaudhary2023code}
Chaudhary, S.
\newblock Code alpaca: An instruction-following llama model for code
  generation.
\newblock \emph{GitHub repository}, 2023.

\bibitem[Chen et~al.(2023)Chen, Li, Yan, Wang, Gunaratna, Yadav, Tang,
  Srinivasan, Zhou, Huang, et~al.]{chen2023alpagasus}
Chen, L., Li, S., Yan, J., Wang, H., Gunaratna, K., Yadav, V., Tang, Z.,
  Srinivasan, V., Zhou, T., Huang, H., et~al.
\newblock Alpagasus: Training a better alpaca with fewer data.
\newblock \emph{arXiv preprint arXiv:2307.08701}, 2023.

\bibitem[Chen et~al.(2021)Chen, Tworek, Jun, Yuan, Pinto, Kaplan, Edwards,
  Burda, Joseph, Brockman, et~al.]{chen2021evaluating}
Chen, M., Tworek, J., Jun, H., Yuan, Q., Pinto, H. P. D.~O., Kaplan, J.,
  Edwards, H., Burda, Y., Joseph, N., Brockman, G., et~al.
\newblock Evaluating large language models trained on code.
\newblock \emph{arXiv preprint arXiv:2107.03374}, 2021.

\bibitem[Clark et~al.(2018)Clark, Cowhey, Etzioni, Khot, Sabharwal, Schoenick,
  and Tafjord]{clark2018think}
Clark, P., Cowhey, I., Etzioni, O., Khot, T., Sabharwal, A., Schoenick, C., and
  Tafjord, O.
\newblock Think you have solved question answering? try arc, the ai2 reasoning
  challenge.
\newblock \emph{arXiv preprint arXiv:1803.05457}, 2018.

\bibitem[Cobbe et~al.(2021)Cobbe, Kosaraju, Bavarian, Chen, Jun, Kaiser,
  Plappert, Tworek, Hilton, Nakano, et~al.]{cobbe2021training}
Cobbe, K., Kosaraju, V., Bavarian, M., Chen, M., Jun, H., Kaiser, L., Plappert,
  M., Tworek, J., Hilton, J., Nakano, R., et~al.
\newblock Training verifiers to solve math word problems.
\newblock \emph{arXiv preprint arXiv:2110.14168}, 2021.

\bibitem[Conover et~al.(2023)Conover, Hayes, Mathur, Xie, Wan, Shah, Ghodsi,
  Wendell, Zaharia, and Xin]{conover2023free}
Conover, M., Hayes, M., Mathur, A., Xie, J., Wan, J., Shah, S., Ghodsi, A.,
  Wendell, P., Zaharia, M., and Xin, R.
\newblock Free dolly: Introducing the world’s first truly open
  instruction-tuned llm.
\newblock \emph{Company Blog of Databricks}, 2023.

\bibitem[Ding et~al.(2023)Ding, Chen, Xu, Qin, Hu, Liu, Sun, and
  Zhou]{ding2023enhancing}
Ding, N., Chen, Y., Xu, B., Qin, Y., Hu, S., Liu, Z., Sun, M., and Zhou, B.
\newblock Enhancing chat language models by scaling high-quality instructional
  conversations.
\newblock In \emph{Proceedings of the 2023 Conference on Empirical Methods in
  Natural Language Processing}, pp.\  3029--3051, 2023.

\bibitem[Dubey et~al.(2024)Dubey, Jauhri, Pandey, Kadian, Al-Dahle, Letman,
  Mathur, Schelten, Yang, Fan, et~al.]{dubey2024llama}
Dubey, A., Jauhri, A., Pandey, A., Kadian, A., Al-Dahle, A., Letman, A.,
  Mathur, A., Schelten, A., Yang, A., Fan, A., et~al.
\newblock The llama 3 herd of models.
\newblock \emph{arXiv preprint arXiv:2407.21783}, 2024.

\bibitem[Hendrycks et~al.()Hendrycks, Burns, Basart, Zou, Mazeika, Song, and
  Steinhardt]{hendrycksmeasuring}
Hendrycks, D., Burns, C., Basart, S., Zou, A., Mazeika, M., Song, D., and
  Steinhardt, J.
\newblock Measuring massive multitask language understanding.
\newblock In \emph{International Conference on Learning Representations}.

\bibitem[Lee et~al.(2024)Lee, Roy, Xu, Raiman, Shoeybi, Catanzaro, and
  Ping]{lee2024nv}
Lee, C., Roy, R., Xu, M., Raiman, J., Shoeybi, M., Catanzaro, B., and Ping, W.
\newblock Nv-embed: Improved techniques for training llms as generalist
  embedding models.
\newblock \emph{arXiv preprint arXiv:2405.17428}, 2024.

\bibitem[Li et~al.(2024{\natexlab{a}})Li, Chen, Chen, He, Gu, and
  Zhou]{li2024selective}
Li, M., Chen, L., Chen, J., He, S., Gu, J., and Zhou, T.
\newblock Selective reflection-tuning: Student-selected data recycling for llm
  instruction-tuning.
\newblock \emph{arXiv preprint arXiv:2402.10110}, 2024{\natexlab{a}}.

\bibitem[Li et~al.(2024{\natexlab{b}})Li, Zhang, He, Li, Zhao, Wang, Cheng, and
  Zhou]{li-etal-2024-superfiltering}
Li, M., Zhang, Y., He, S., Li, Z., Zhao, H., Wang, J., Cheng, N., and Zhou, T.
\newblock Superfiltering: Weak-to-strong data filtering for fast
  instruction-tuning.
\newblock In Ku, L.-W., Martins, A., and Srikumar, V. (eds.), \emph{Proceedings
  of the 62nd Annual Meeting of the Association for Computational Linguistics
  (Volume 1: Long Papers)}, pp.\  14255--14273, 2024{\natexlab{b}}.
\newblock \doi{10.18653/v1/2024.acl-long.769}.

\bibitem[Li et~al.(2024{\natexlab{c}})Li, Zhang, Li, Chen, Chen, Cheng, Wang,
  Zhou, and Xiao]{li2024quantity}
Li, M., Zhang, Y., Li, Z., Chen, J., Chen, L., Cheng, N., Wang, J., Zhou, T.,
  and Xiao, J.
\newblock From quantity to quality: Boosting llm performance with self-guided
  data selection for instruction tuning.
\newblock In \emph{Proceedings of the 2024 Conference of the North American
  Chapter of the Association for Computational Linguistics: Human Language
  Technologies (Volume 1: Long Papers)}, pp.\  7595--7628, 2024{\natexlab{c}}.

\bibitem[Li et~al.(2024{\natexlab{d}})Li, Hui, Xia, Yang, Yang, Zhang, Si,
  Chen, Liu, Liu, Huang, and Li]{li-etal-2024-one}
Li, Y., Hui, B., Xia, X., Yang, J., Yang, M., Zhang, L., Si, S., Chen, L.-H.,
  Liu, J., Liu, T., Huang, F., and Li, Y.
\newblock One-shot learning as instruction data prospector for large language
  models.
\newblock In Ku, L.-W., Martins, A., and Srikumar, V. (eds.), \emph{Proceedings
  of the 62nd Annual Meeting of the Association for Computational Linguistics
  (Volume 1: Long Papers)}, pp.\  4586--4601, 2024{\natexlab{d}}.
\newblock \doi{10.18653/v1/2024.acl-long.252}.

\bibitem[Lin et~al.(2022)Lin, Hilton, and Evans]{lin2022truthfulqa}
Lin, S., Hilton, J., and Evans, O.
\newblock Truthfulqa: Measuring how models mimic human falsehoods.
\newblock In \emph{Proceedings of the 60th Annual Meeting of the Association
  for Computational Linguistics (Volume 1: Long Papers)}, pp.\  3214--3252,
  2022.

\bibitem[Liu et~al.(2023)Liu, Xia, Wang, and Zhang]{evalplus}
Liu, J., Xia, C.~S., Wang, Y., and Zhang, L.
\newblock Is your code generated by chat{GPT} really correct? rigorous
  evaluation of large language models for code generation.
\newblock In \emph{Thirty-seventh Conference on Neural Information Processing
  Systems}, 2023.
\newblock URL \url{https://openreview.net/forum?id=1qvx610Cu7}.

\bibitem[Liu et~al.(2024{\natexlab{a}})Liu, Liu, Wong, Li, Wang, Hu, and
  Zhang]{liu2024selectit}
Liu, L., Liu, X., Wong, D.~F., Li, D., Wang, Z., Hu, B., and Zhang, M.
\newblock Selectit: Selective instruction tuning for large language models via
  uncertainty-aware self-reflection.
\newblock \emph{arXiv preprint arXiv:2402.16705}, 2024{\natexlab{a}}.

\bibitem[Liu et~al.(2024{\natexlab{b}})Liu, Zeng, He, Jiang, and He]{liumakes}
Liu, W., Zeng, W., He, K., Jiang, Y., and He, J.
\newblock What makes good data for alignment? a comprehensive study of
  automatic data selection in instruction tuning.
\newblock In \emph{The Twelfth International Conference on Learning
  Representations}, 2024{\natexlab{b}}.

\bibitem[Longpre et~al.(2023)Longpre, Hou, Vu, Webson, Chung, Tay, Zhou, Le,
  Zoph, Wei, et~al.]{longpre2023flan}
Longpre, S., Hou, L., Vu, T., Webson, A., Chung, H.~W., Tay, Y., Zhou, D., Le,
  Q.~V., Zoph, B., Wei, J., et~al.
\newblock The flan collection: Designing data and methods for effective
  instruction tuning.
\newblock In \emph{International Conference on Machine Learning}, pp.\
  22631--22648. PMLR, 2023.

\bibitem[Lu et~al.(2023)Lu, Yuan, Yuan, Lin, Lin, Tan, Zhou, and
  Zhou]{lu2023instag}
Lu, K., Yuan, H., Yuan, Z., Lin, R., Lin, J., Tan, C., Zhou, C., and Zhou, J.
\newblock \# instag: Instruction tagging for analyzing supervised fine-tuning
  of large language models.
\newblock In \emph{The Twelfth International Conference on Learning
  Representations}, 2023.

\bibitem[Muennighoff et~al.(2022)Muennighoff, Tazi, Magne, and
  Reimers]{muennighoff2022mteb}
Muennighoff, N., Tazi, N., Magne, L., and Reimers, N.
\newblock Mteb: Massive text embedding benchmark.
\newblock \emph{arXiv preprint arXiv:2210.07316}, 2022.
\newblock \doi{10.48550/ARXIV.2210.07316}.
\newblock URL \url{https://arxiv.org/abs/2210.07316}.

\bibitem[Ni et~al.(2024)Ni, Xue, Yue, Deng, Shah, Jain, Neubig, and
  You]{ni2024mixeval}
Ni, J., Xue, F., Yue, X., Deng, Y., Shah, M., Jain, K., Neubig, G., and You, Y.
\newblock Mixeval: Deriving wisdom of the crowd from llm benchmark mixtures.
\newblock \emph{arXiv preprint arXiv:2406.06565}, 2024.

\bibitem[Ouyang et~al.(2022)Ouyang, Wu, Jiang, Almeida, Wainwright, Mishkin,
  Zhang, Agarwal, Slama, Ray, et~al.]{ouyang2022training}
Ouyang, L., Wu, J., Jiang, X., Almeida, D., Wainwright, C., Mishkin, P., Zhang,
  C., Agarwal, S., Slama, K., Ray, A., et~al.
\newblock Training language models to follow instructions with human feedback.
\newblock \emph{Advances in neural information processing systems},
  35:\penalty0 27730--27744, 2022.

\bibitem[Peng et~al.(2023)Peng, Li, He, Galley, and Gao]{peng2023instruction}
Peng, B., Li, C., He, P., Galley, M., and Gao, J.
\newblock Instruction tuning with gpt-4.
\newblock \emph{arXiv preprint arXiv:2304.03277}, 2023.

\bibitem[Radford et~al.(2019)Radford, Wu, Child, Luan, Amodei, Sutskever,
  et~al.]{radford2019language}
Radford, A., Wu, J., Child, R., Luan, D., Amodei, D., Sutskever, I., et~al.
\newblock Language models are unsupervised multitask learners.
\newblock \emph{OpenAI blog}, 1\penalty0 (8):\penalty0 9, 2019.

\bibitem[Shen(2024)]{shen2024rethinking}
Shen, M.
\newblock Rethinking data selection for supervised fine-tuning.
\newblock \emph{arXiv preprint arXiv:2402.06094}, 2024.

\bibitem[Suzgun et~al.(2023)Suzgun, Scales, Sch{\"a}rli, Gehrmann, Tay, Chung,
  Chowdhery, Le, Chi, Zhou, et~al.]{suzgun2023challenging}
Suzgun, M., Scales, N., Sch{\"a}rli, N., Gehrmann, S., Tay, Y., Chung, H.~W.,
  Chowdhery, A., Le, Q., Chi, E., Zhou, D., et~al.
\newblock Challenging big-bench tasks and whether chain-of-thought can solve
  them.
\newblock In \emph{Findings of the Association for Computational Linguistics:
  ACL 2023}, pp.\  13003--13051, 2023.

\bibitem[Taori et~al.(2023)Taori, Gulrajani, Zhang, Dubois, Li, Guestrin,
  Liang, and Hashimoto]{alpaca}
Taori, R., Gulrajani, I., Zhang, T., Dubois, Y., Li, X., Guestrin, C., Liang,
  P., and Hashimoto, T.~B.
\newblock Stanford alpaca: An instruction-following llama model.
\newblock \url{https://github.com/tatsu-lab/stanford_alpaca}, 2023.

\bibitem[Team(2024)]{qwen25}
Team, Q.
\newblock Qwen2.5: A party of foundation models, September 2024.
\newblock URL \url{https://qwenlm.github.io/blog/qwen2.5/}.

\bibitem[Wang et~al.(2024)Wang, Xiong, Xie, Zhao, and
  Zhang]{wang2024interpretable}
Wang, H., Xiong, W., Xie, T., Zhao, H., and Zhang, T.
\newblock Interpretable preferences via multi-objective reward modeling and
  mixture-of-experts.
\newblock \emph{arXiv preprint arXiv:2406.12845}, 2024.

\bibitem[Wang et~al.(2022)Wang, Mishra, Alipoormolabashi, Kordi, Mirzaei,
  Arunkumar, Ashok, Dhanasekaran, Naik, Stap, et~al.]{wang2022super}
Wang, Y., Mishra, S., Alipoormolabashi, P., Kordi, Y., Mirzaei, A., Arunkumar,
  A., Ashok, A., Dhanasekaran, A.~S., Naik, A., Stap, D., et~al.
\newblock Super-naturalinstructions: Generalization via declarative
  instructions on 1600+ nlp tasks.
\newblock \emph{arXiv preprint arXiv:2204.07705}, 2022.

\bibitem[Wang et~al.(2023)Wang, Kordi, Mishra, Liu, Smith, Khashabi, and
  Hajishirzi]{wang2023self}
Wang, Y., Kordi, Y., Mishra, S., Liu, A., Smith, N.~A., Khashabi, D., and
  Hajishirzi, H.
\newblock Self-instruct: Aligning language models with self-generated
  instructions.
\newblock In \emph{Proceedings of the 61st Annual Meeting of the Association
  for Computational Linguistics (Volume 1: Long Papers)}, pp.\  13484--13508,
  2023.

\bibitem[Wei et~al.(2022)Wei, Bosma, Zhao, Guu, Yu, Lester, Du, Dai, and
  Le]{weifinetuned}
Wei, J., Bosma, M., Zhao, V., Guu, K., Yu, A.~W., Lester, B., Du, N., Dai,
  A.~M., and Le, Q.~V.
\newblock Finetuned language models are zero-shot learners.
\newblock In \emph{International Conference on Learning Representations}, 2022.

\bibitem[Wu et~al.(2024{\natexlab{a}})Wu, Vu, Qu, and Haffari]{wu2024best}
Wu, M., Vu, T.-T., Qu, L., and Haffari, G.
\newblock The best of both worlds: Bridging quality and diversity in data
  selection with bipartite graph.
\newblock \emph{arXiv preprint arXiv:2410.12458}, 2024{\natexlab{a}}.

\bibitem[Wu et~al.(2024{\natexlab{b}})Wu, Waheed, Zhang, Abdul-Mageed, and
  Aji]{wu2024lamini}
Wu, M., Waheed, A., Zhang, C., Abdul-Mageed, M., and Aji, A.
\newblock Lamini-lm: A diverse herd of distilled models from large-scale
  instructions.
\newblock In \emph{Proceedings of the 18th Conference of the European Chapter
  of the Association for Computational Linguistics (Volume 1: Long Papers)},
  pp.\  944--964, 2024{\natexlab{b}}.

\bibitem[Xu et~al.(2023)Xu, Sun, Zheng, Geng, Zhao, Feng, Tao, and
  Jiang]{xu2023wizardlm}
Xu, C., Sun, Q., Zheng, K., Geng, X., Zhao, P., Feng, J., Tao, C., and Jiang,
  D.
\newblock Wizardlm: Empowering large language models to follow complex
  instructions.
\newblock \emph{arXiv preprint arXiv:2304.12244}, 2023.

\bibitem[Zellers et~al.(2019)Zellers, Holtzman, Bisk, Farhadi, and
  Choi]{zellers2019hellaswag}
Zellers, R., Holtzman, A., Bisk, Y., Farhadi, A., and Choi, Y.
\newblock Hellaswag: Can a machine really finish your sentence?
\newblock In \emph{Proceedings of the 57th Annual Meeting of the Association
  for Computational Linguistics}, pp.\  4791--4800, 2019.

\bibitem[Zhao et~al.(2024)Zhao, Andriushchenko, Croce, and
  Flammarion]{zhao2024long}
Zhao, H., Andriushchenko, M., Croce, F., and Flammarion, N.
\newblock Long is more for alignment: A simple but tough-to-beat baseline for
  instruction fine-tuning.
\newblock In \emph{Forty-first International Conference on Machine Learning},
  2024.

\bibitem[Zhou et~al.(2024)Zhou, Liu, Xu, Iyer, Sun, Mao, Ma, Efrat, Yu, Yu,
  et~al.]{zhou2024lima}
Zhou, C., Liu, P., Xu, P., Iyer, S., Sun, J., Mao, Y., Ma, X., Efrat, A., Yu,
  P., Yu, L., et~al.
\newblock Lima: Less is more for alignment.
\newblock \emph{Advances in Neural Information Processing Systems}, 36, 2024.

\end{thebibliography}
\bibliographystyle{icml2025}

\newpage
\appendix
\appendix

\section{Performance on Dolly Dataset}
\label{app:dolly-results}

The performance of Dolly is shown in Table~\ref{tab:dolly-results}. All of the approaches struggle to make improvements in all aspects. GraphFilter achieves superior scores on 4 out of 7 benchmarks, with a compromise on math reasoning ability, leading to its inferior overall performance. Longest and \ApproachName{} ranks top 1 and 2 according to the average score, respectively, with \ApproachName{} topping the rank on MixEval.

\begin{table*}[h]
    \centering
    \caption{Performance of approaches backed on Llama-3-8B fine-tuned with Dolly. The highest score in each column is in bold, and the second-best ones for overall performance are underlined. $\star$ marks our proposed approach.}
    \scalebox{1.0}{\begin{tabular}{l|ccccccc|c|c}
        \toprule[1pt]
         Method & GSM8K & MMLU & \makecell{Truthful\\QA} & BBH & \makecell{Human\\Eval} & ARC & \makecell{Hella\\Swag} & AVG & MixEval\\
        \midrule[1pt]
        \multicolumn{10}{l}{\textit{Results on Dolly}} \\
        Vanilla & 49.5 & 56.14 & 42.51 & 55.65 & 42.32 & 60.75 & 83.78 & 55.81 & 28.70\\
        Longest & 63.00 & \textbf{61.09} & 42.88 & 59.54 & 41.34 & 61.43 & 84.72 & \textbf{59.14} & {32.55} \\
        Deita & 48.00 & 57.04 & 37.01 & 53.33 & 42.80 & 63.05 & 84.22 & 55.06 & 28.85 \\
        Superfiltering & 57.50 & 60.13 & 42.52 & 58.43 & 40.97 & \textbf{63.57} & 84.81 & 58.28 & 31.50 \\
        GraphFilter & 53.00 & 60.77 & \textbf{43.67} & \textbf{59.63} & \textbf{44.76} & 61.86 & \textbf{84.89} & 58.36 & \underline{32.85} \\
        \ApproachName{}$\star$ & \textbf{65.00} & 60.70 & 41.92 & 56.20 & 41.71 & 61.60 & 84.64 & \underline{58.82} & \textbf{33.80} \\
        \bottomrule[1pt]
    \end{tabular}}
    \label{tab:dolly-results}
\end{table*}

\section{Performance on Qwen-2.5-7B}
\label{app:qwen-results}

Detailed performance on Qwen-2.5-7B is in Table~\ref{tab:qwen-results-all}. Our approach \ApproachName{}'s overall performance consistently outperforms Vanilla, and ranks toppest on 5 out of 6 cases.

\begin{table*}[h]
    \centering
    \caption{Performance of approaches backed on Qwen-2.5-7B.}
    \scalebox{1.0}{\begin{tabular}{l|ccccccc|c|c}
        \toprule[1pt]
         Method & GSM8K & MMLU & \makecell{Truthful\\QA} & BBH & \makecell{Human\\Eval} & ARC & \makecell{Hella\\Swag} & AVG & MixEval\\
        \midrule[1pt]
        \multicolumn{10}{l}{\textit{Results on Alpaca}} \\
        Vanilla & 74.00 & 69.50 & \textbf{46.50} & \textbf{66.94} & 54.27 & \textbf{64.85} & \textbf{80.14} & 65.17 & 38.30 \\
        Longest &  \textbf{86.00} & 70.74 & 44.95 & 66.20 & 53.78 & 63.91 & 79.90 & \textbf{66.49} & \underline{38.45} \\
        \ApproachName{}$\star$ & 84.00 & \textbf{70.77} & 45.32 & \textbf{66.94} & \textbf{54.76} & 63.31 & 79.94 & \underline{66.43} & \textbf{38.75} \\
        \midrule[1pt]
        \multicolumn{10}{l}{\textit{Results on Alpaca-GPT4}} \\
        Vanilla & 85.50 & 70.91 & 55.30 & 66.30 & 62.07 & 63.23 & \textbf{80.85} & \underline{69.17} & 38.20 \\
        Longest & 87.00 & 71.38 & 56.75 & 65.28 & \textbf{58.78} & 63.57 & 80.59 & 69.05 & \underline{39.50} \\
        \ApproachName{}$\star$ & \textbf{87.50} & \textbf{71.41} & \textbf{57.22} & \textbf{66.67} & 57.93 & \textbf{63.82} & 80.68 & \textbf{69.32} & \textbf{40.50} \\
        \midrule[1pt]
        \multicolumn{10}{l}{\textit{Results on WizardLM}} \\
        Vanilla & \textbf{88.50} & 70.62 & \textbf{59.44} & \textbf{65.37} & 56.10 & 62.20 & 80.51 & \underline{68.96} & \underline{40.15}\\
        Longest & 87.50 & \textbf{71.31} & 56.46 & 62.69 & 56.59 & 64.68 & 81.36 & 68.66 & 39.50 \\
        \ApproachName{}$\star$ & 88.00 & 71.14 & 57.27 & 63.43 & \textbf{56.95} & \textbf{65.02} & \textbf{81.50} & \textbf{69.04} & \textbf{40.95} \\
        \bottomrule[1pt]
    \end{tabular}}
    \label{tab:qwen-results-all}
\end{table*}

\section{Analysis of Selected Data}
\label{app:response-length}

The statistics of response lengths for instruction-tuning data selected from Alpaca-GPT4 and WizardLM by different approaches are depicted in Figure~\ref{fig:length-analysis-2}. The average response length of data chosen by our \ApproachName{} is longer than the other baselines, while covering data with short responses compared to Longest. 

\begin{figure*}[h!]
    \centering
    \subfigure[\small{Alpaca-GPT4}\label{fig:length-alpaca-gpt4}]{
    \begin{minipage}[t]{0.5\linewidth}
    \centering
    \includegraphics[width=\linewidth]{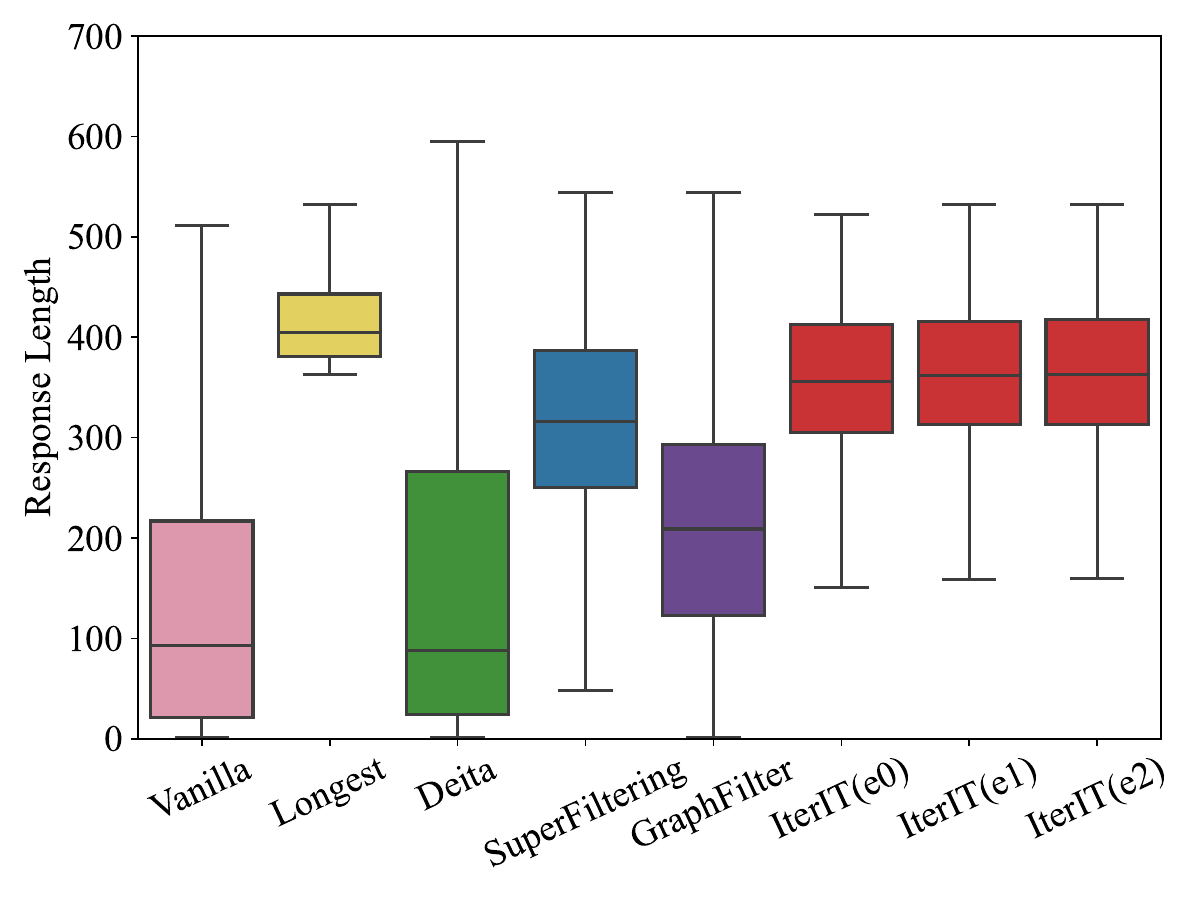}
    \end{minipage}%
    }%
    \subfigure[\small WizardLM\label{fig:length-wizard}]{
    \begin{minipage}[t]{0.5\linewidth}
    \centering
    \includegraphics[width=\linewidth]{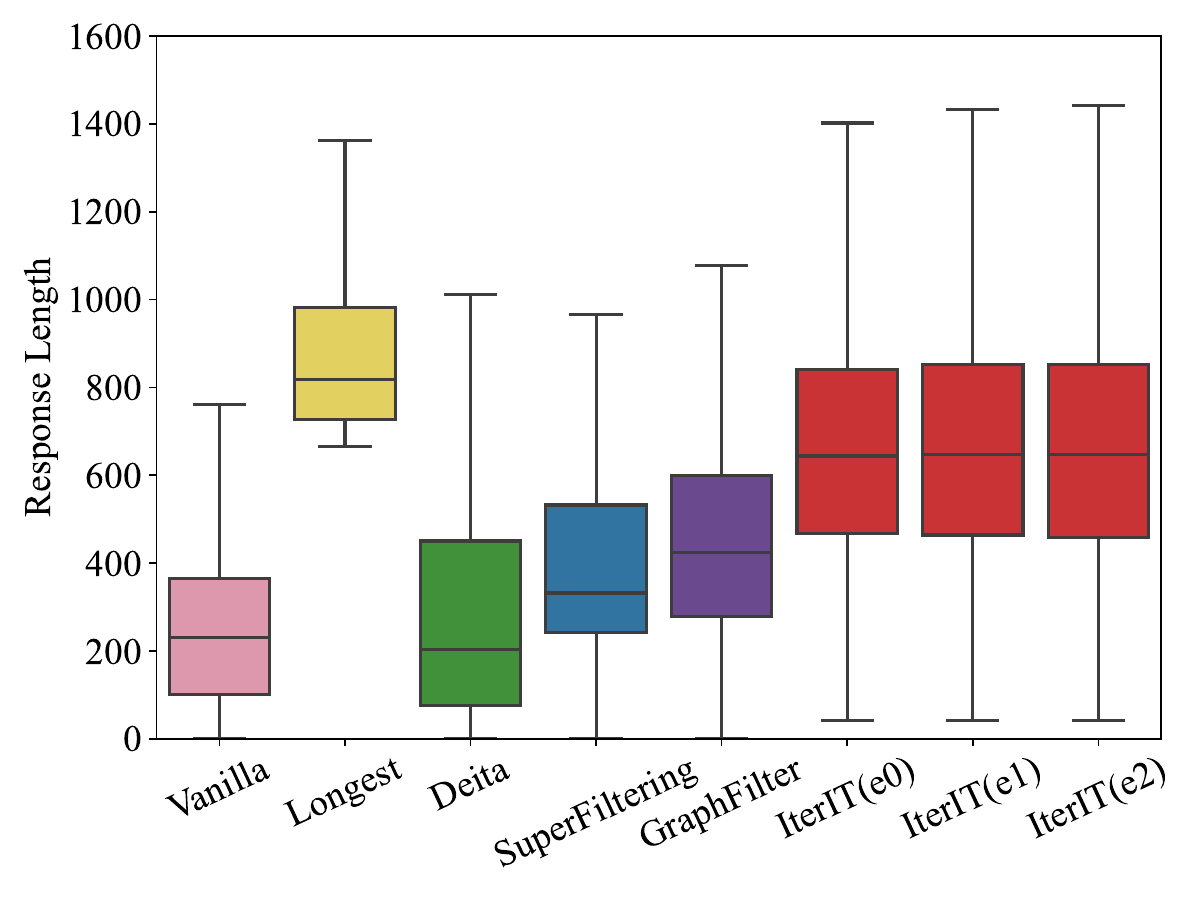}
    \end{minipage}%
    }%
    \caption{Response lengths of instruction-tuning data selected from Alpaca-GPT4 and WizardLM.}
    \label{fig:length-analysis-2}
\end{figure*}

The Jaccard similarity between the subsets selected before each epoch on Alpaca-GPT4 and WizardLM is in Table~\ref{tab:jaccard-sim-all}. The trends are identical to that of Alpaca.
\begin{table}[]
    \centering
    \caption{Jaccard similarity(\%) between pairs of subset selected by \ApproachName{} during instruction tuning.}
    \begin{tabular}{c|ccc}
    \toprule[1pt]
         Epochs& Alpaca-GPT4 & WizardLM  \\
    \midrule[1pt]
         1,2 & 81.06 & 83.29 \\
         2,3 & 82.20 & 85.48 \\
         1,3 & 73.04 & 77.48 \\ 
    \bottomrule[1pt]
    \end{tabular}
    \label{tab:jaccard-sim-all}
\end{table}

\begin{figure*}[h!]
    \centering
    
    \subfigure[\small Longest\label{fig:tsne-longest}]{
    \begin{minipage}[t]{0.33\linewidth}
    \centering
    \includegraphics[width=\linewidth]{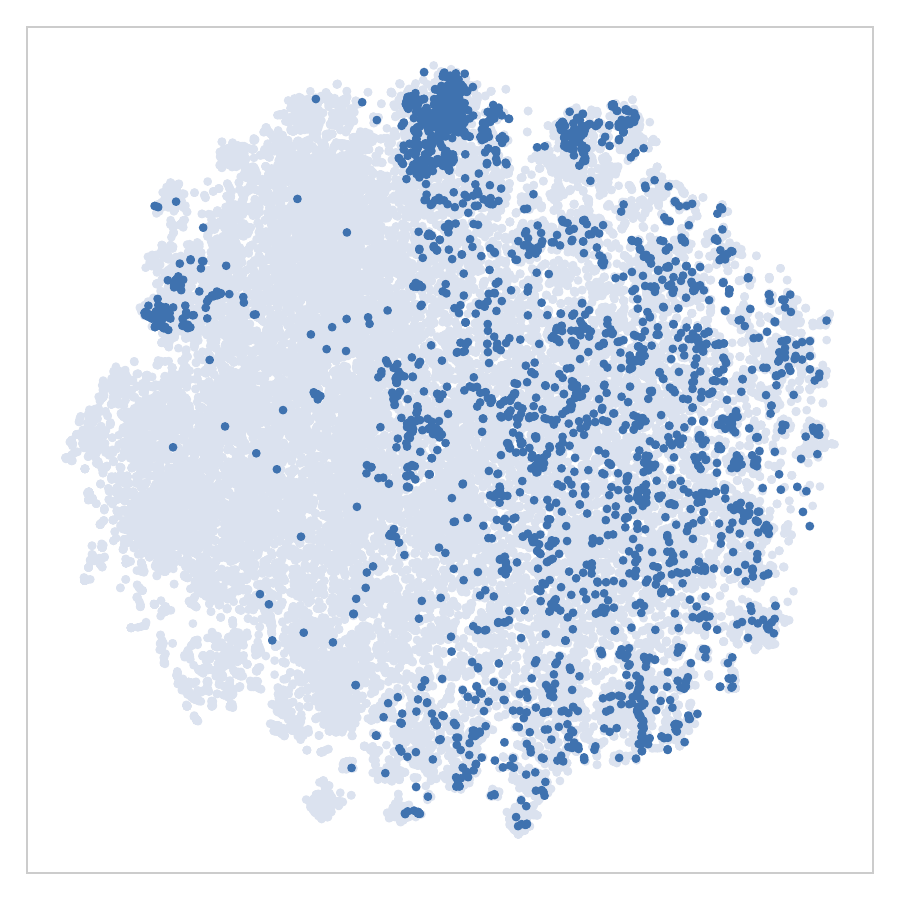}
    \end{minipage}%
    }%
    \subfigure[\small{Deita}\label{fig:tsne-deita}]{
    \begin{minipage}[t]{0.33\linewidth}
    \centering
    \includegraphics[width=\linewidth]{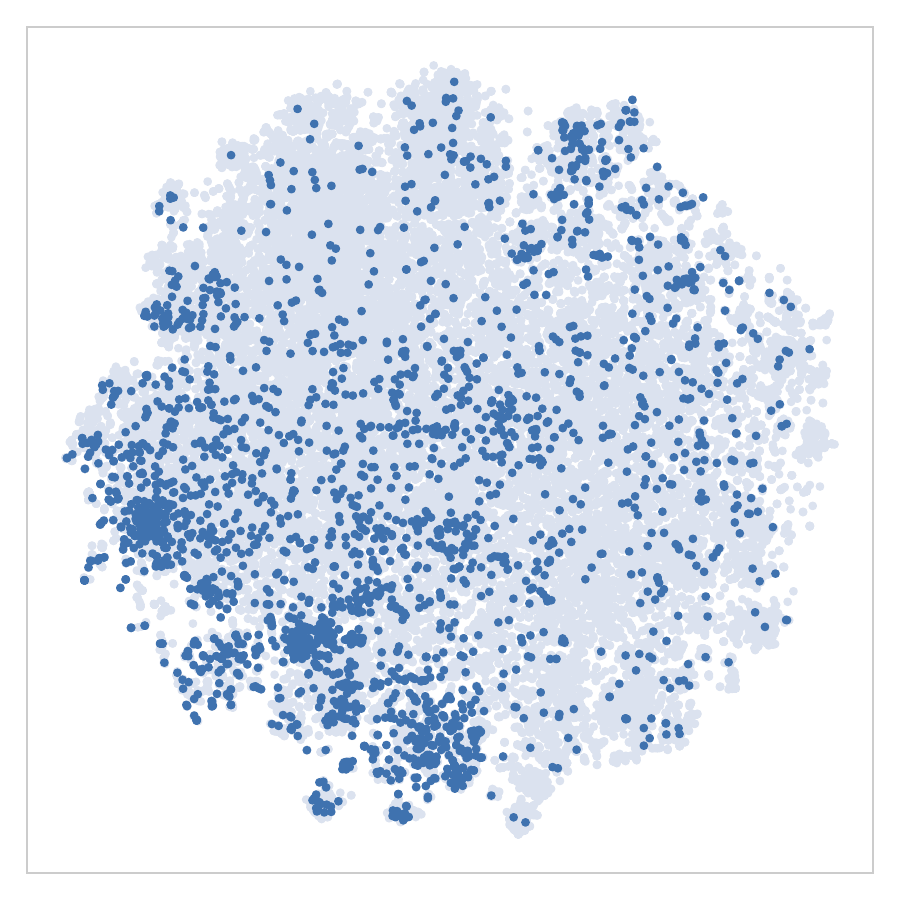}
    \end{minipage}%
    }%
    \subfigure[\small Superfiltering\label{fig:tsne-superfiltering}]{
    \begin{minipage}[t]{0.33\linewidth}
    \centering
    \includegraphics[width=\linewidth]{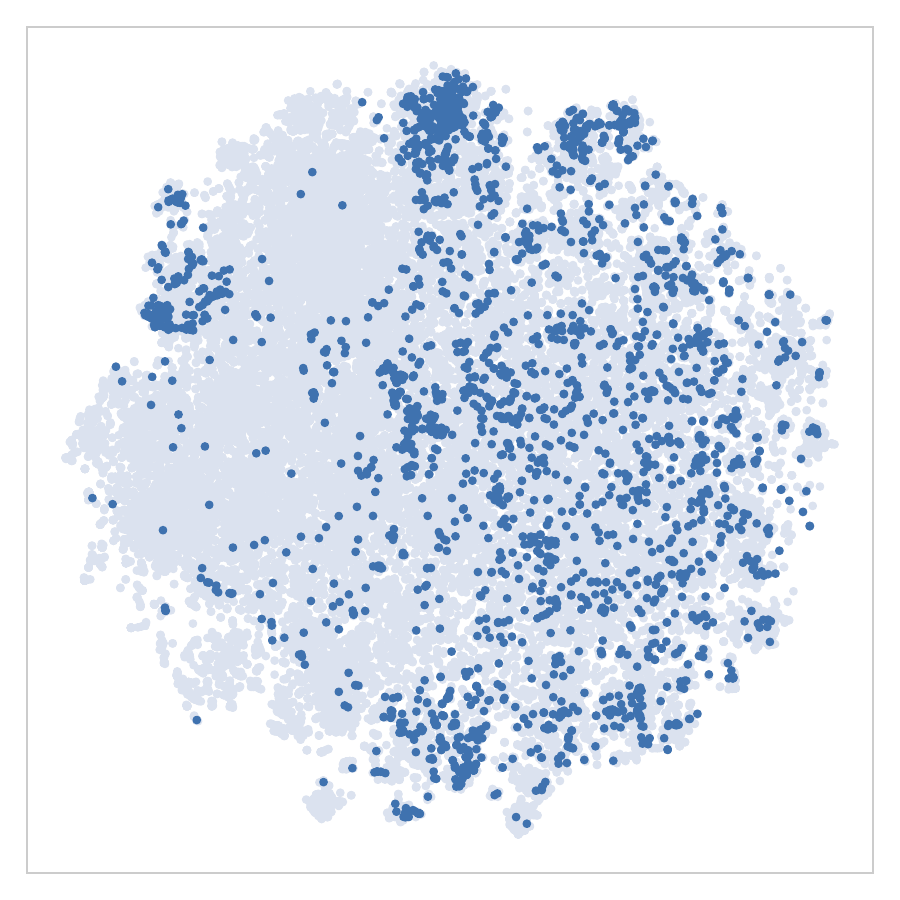}
    \end{minipage}%
    }%

    \subfigure[\small GraphFilter\label{fig:tsne-graphfilter}]{
    \begin{minipage}[t]{0.33\linewidth}
    \centering
    \includegraphics[width=\linewidth]{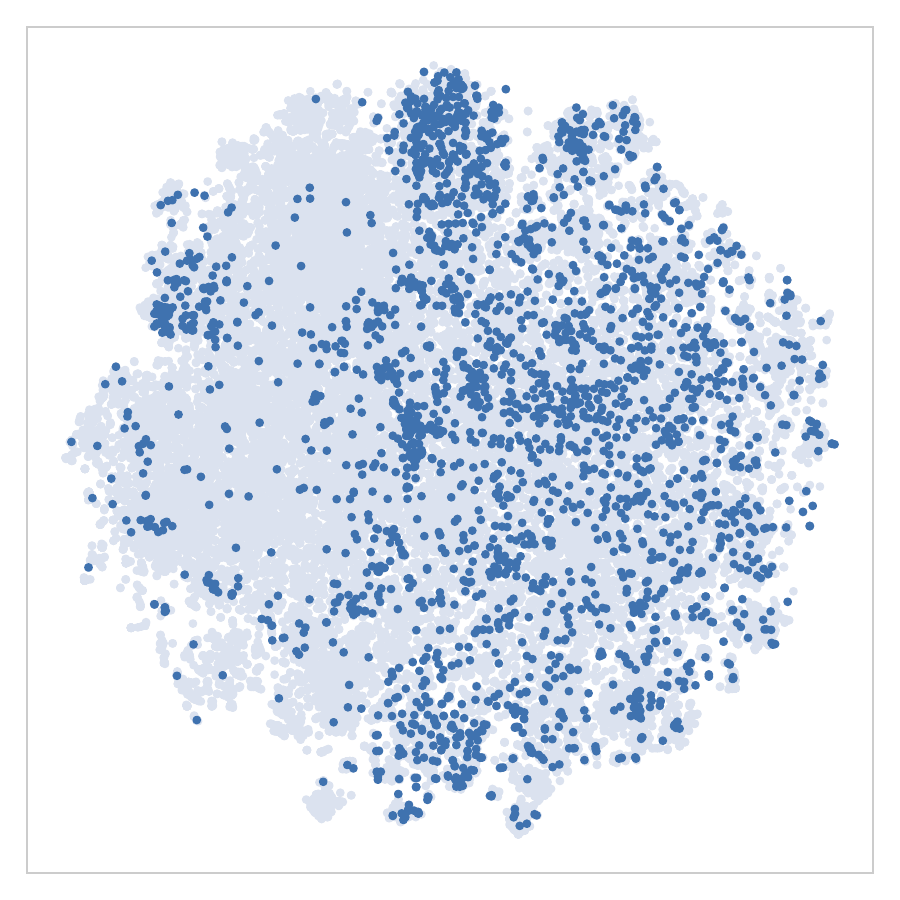}
    \end{minipage}%
    }%
    \subfigure[\small{\ApproachName{}}\label{fig:tsne-our}]{
    \begin{minipage}[t]{0.33\linewidth}
    \centering
    \includegraphics[width=\linewidth]{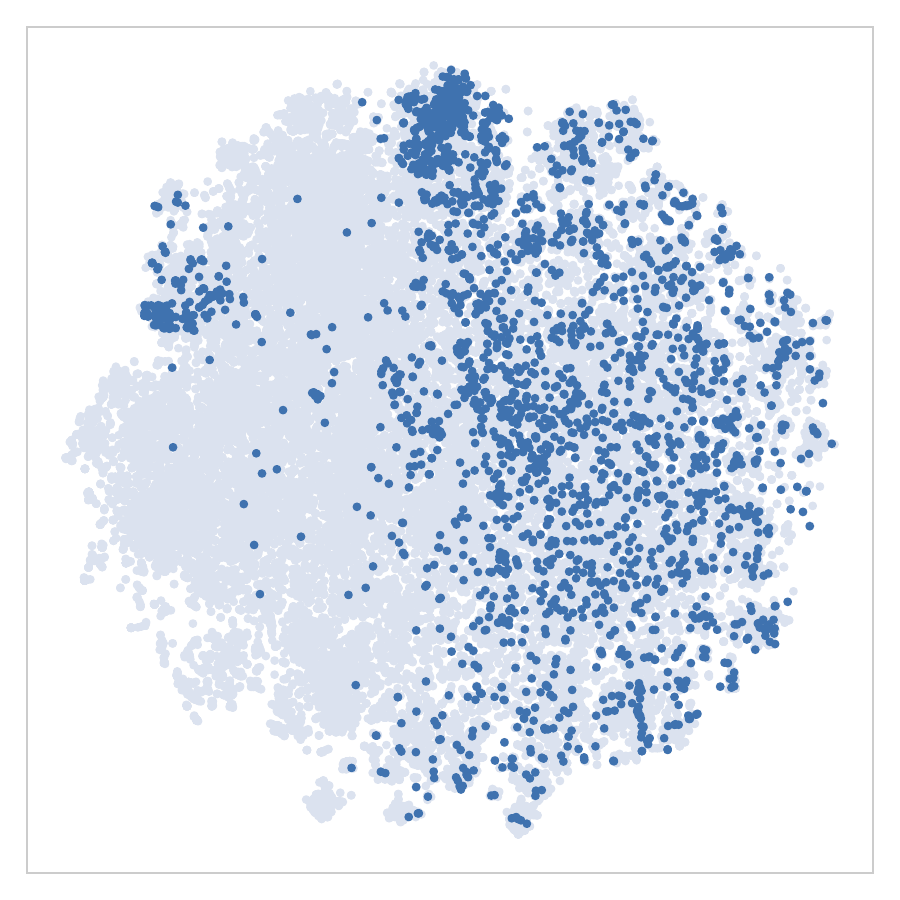}
    \end{minipage}%
    }%
    \caption{Visualization using t-SNE of data points from Alpaca dataset. Grey points represent samples from the dataset, and blue ones represent samples selected by corresponding approaches.}
    \label{fig:scatter-plot}
\end{figure*}

We further extract the response representations of samples by NV-Embed~\cite{lee2024nv}, which achieves the state-of-the-art performance on MTEB benchmark~\cite{muennighoff2022mteb}. The scatter plot visualized using t-SNE on response representations from the Alpaca dataset is shown in Figure~\ref{fig:scatter-plot}. Deita and GraphFilter select data with better semantic diversity, while both of them don't reflect superior performance in Table~\ref{tab:main-results}. \ApproachName{} shares some semantic distribution similarities with Longest. Nonetheless, the specific data points selected by both approaches are different as analyzed in Sec.~\ref{sec:analysis-selected-data}. \ApproachName{} shows favorable performance and strong generalization ability, demonstrating its successful design of combining the complexity and diversity metrics based on the instruction-following difficulty and response informativeness, respectively.

\end{document}